\documentclass[10pt,journal,compsoc]{IEEEtran}

\usepackage{amsmath,amsfonts}
\usepackage{algorithmic}
\usepackage{booktabs}
\usepackage{algorithm}
\usepackage{array}
\usepackage[caption=false,font=normalsize,labelfont=sf,textfont=sf]{subfig}
\usepackage{textcomp}
\usepackage{stfloats}
\usepackage{url}
\usepackage{verbatim}
\usepackage{arydshln}
\usepackage{graphicx}
\usepackage[numbers,sort&compress]{natbib}
\usepackage{multirow}
\usepackage{xspace}
\usepackage{booktabs}
\usepackage{makecell}
\renewcommand{\arraystretch}{1.3}
\usepackage{xcolor}

\newcommand{\etc}{\emph{etc.}\xspace}

\newcommand{\ie}{\emph{i.e.,}\xspace}
\usepackage[most,breakable]{tcolorbox}
\usepackage{amsthm}

\newtcolorbox{mytheorem}{
  colback=gray!5,
  colframe=gray!80, 
  boxrule=0.5pt, 
  arc=4pt, 
  left=4pt, 
  right=4pt, 
  top=4pt, 
  bottom=4pt, 
}
\usetikzlibrary{calc}
\usetikzlibrary{positioning,shapes.misc}

\tikzset{
  pics/phase/.style args={#1,#2,#3}{
      code={
          \node[] (x1) at (0, 0) {
            \begin{minipage}[c]{6.8cm}
              \small{#2}
            \end{minipage}
          };

          \draw[rounded corners,draw=#3]  (x1.north west) -- (x1.north east) -- (x1.south east) -- (x1.south west);

          \draw[fill=#3,draw=none]
          (x1.north west) --
          (x1.south west) --
          ($(x1.south west)+(-0.75,-0.25)$) --
          ($(x1.south west)+(-1.5,0)$) --
          ($(x1.north west)+(-1.5,0)$) --
          ($(x1.north west)+(-0.75,-0.25)$) -- cycle;

          \node[rectangle,draw,text=white] at ($(x1.west)+(-0.75,-0.2)$) {
            \begin{minipage}[c]{3.4em}
              \centering
              \small{#1}
            \end{minipage}
          };
        }
    },
}

\NewDocumentCommand{\heng}
{ mO{} }{\textcolor{red}{\textsuperscript{\textit{Heng}}\textsf{\textbf{\small[#1]}}}}

\begin{document}

\title{Toward Generalizable Evaluation in the LLM Era: A Survey Beyond Benchmarks}

\author{Yixin Cao$^1$, Shibo Hong$^1$, Xinze Li$^2$, Jiahao Ying$^3$, Yubo Ma$^2$, Haiyuan Liang$^1$, Yantao Liu$^1$, Zijun Yao$^4$, Xiaozhi Wang$^4$, Dan Huang$^3$, Wenxuan Zhang$^5$, Lifu Huang$^6$, Muhao Chen$^6$, Lei Hou$^4$, Qianru Sun$^3$, Xingjun Ma$^1$, Zuxuan Wu$^1$, Min-Yen Kan$^7$, David Lo$^3$, Qi Zhang$^1$, Heng Ji$^8$, Jing Jiang$^9$, Juanzi Li$^4$, Aixin Sun$^2$, Xuanjing Huang$^1$, Tat-Seng Chua$^7$, Yu-Gang Jiang$^1$ \\

$^1$Fudan University, $^2$Nanyang Technological University, $^3$Singapore Management University, $^4$Tsinghua University, $^5$Singapore University of Technology and Design, $^6$University of California Davis, $^7$National University of Singapore, $^8$University of Illinois Urbana-Champaign, $^9$Australian National University
\thanks{Corresponding author: Yixin Cao, yxcao@fudan.edu.cn}
}

\markboth{Cao \MakeLowercase{\textit{et al.}}: Toward Generalizable Evaluation in the LLM Era: A Survey Beyond Benchmarks}{April 2025}

\IEEEtitleabstractindextext{%

\begin{abstract}
Large Language Models (LLMs) are advancing at an amazing speed and have become indispensable across academia, industry, and daily applications. To keep pace with the status quo, this survey probes the core challenges that the rise of LLMs poses for evaluation. We identify and analyze two pivotal transitions: (i) from task‑specific to capability‑based evaluation, which reorganizes benchmarks around core competencies such as knowledge, reasoning, instruction following, multi‑modal understanding, and safety; and (ii) from manual to automated evaluation, encompassing dynamic dataset curation and ``LLM‑as‑a‑judge'' scoring.

Yet, even with these transitions, a crucial obstacle persists: the evaluation generalization issue. Bounded test sets cannot scale alongside models whose abilities grow seemingly without limit. We will dissect this issue, along with the core challenges of the above two transitions, from the perspectives of methods, datasets, evaluators, and metrics.
Due to the fast evolving of this field, we will maintain a living GitHub repository (links are in each section) to crowd‑source updates and corrections, and warmly invite contributors and collaborators.
\end{abstract}

}

\maketitle

\IEEEdisplaynontitleabstractindextext

\section{Introduction}
\label{sec:intro}

Large Language Models (LLMs) have achieved unprecedented success in both academia and industry, largely attributed to the rapid advancements in training and evaluation techniques. As the ``quality‑control system'', evaluation not only guides the trajectory of technological progress but also serves as an early-warning mechanism for potential risks. Recent reasoning LLMs like OpenAI o1 or DeepSeek‑R1 further underscore this importance of evaluation --- by integrating reasoning, evaluation, and subsequent re‑reasoning (i.e., refinement or correction) into a single Chain‑of‑Thought (CoT), their inference quality got greatly improved.
These advances have invigorated the evaluation community, producing an ever‑expanding array of benchmarks and assessment studies.
To keep pace with this rapid growth, our survey goes beyond mere cataloging or facet‑specific reviews. Instead, we delve into the fundamental challenges by examining how the advent of LLMs has reshaped the evaluation landscape, a phenomenon we term the \textbf{evaluation generalization}.

\begin{figure*}
\centering
\includegraphics[width=\textwidth]{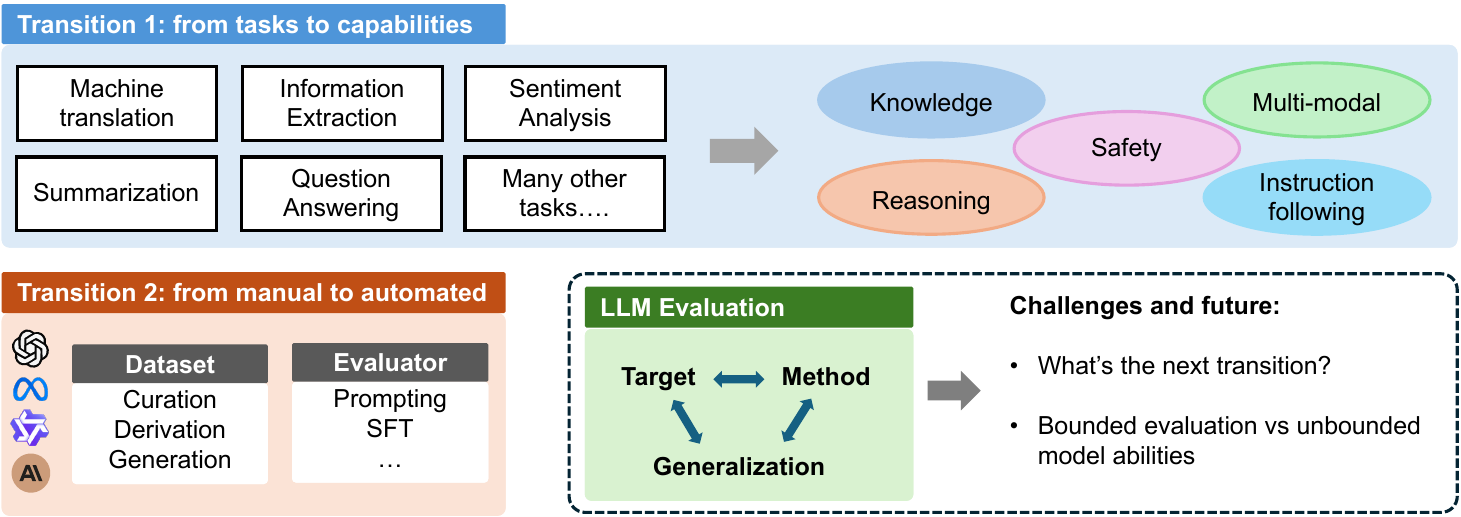}
\caption{Illustration of two transitions in the field of evaluation for LLMs.}
\label{fig:transition}
\end{figure*}

Upon reviewing current research in this area, we identify two critical transitions. As shown in Figure~\ref{fig:transition}, one transition in evaluation is \textit{from task-specific to capability-based}. Traditional evaluation methods focused on specific tasks (e.g., text classification, information extraction). As LLMs unify various NLP tasks in the same form of natural language generation, the definition of each task and the boundaries between them has become increasingly blurred. In this new paradigm, each instruction or prompt can be viewed as an individual task, shifting
attention toward assessing the core capabilities needed to tackle real-world needs. In this survey, we identify five key capabilities: knowledge, reasoning, instruction following\footnote{Conventional NLP tasks are considered part of instruction following.}, multi-modal understanding, and safety. In Section~\ref{sec:bench}, we survey existing benchmarks and categorize them within this capability framework, further dividing them into more detailed sub-categories. In addition, we discuss comprehensive evaluations that assess the interplay between different capabilities and current live leaderboards. This shift from task-based to capability-based evaluation enables a comprehensive understanding of a model’s true potential, beyond its performance in predefined tasks.

Another transition in evaluation is \textit{from manual to automated} methods, including data curation and judgment.
On the data side, rapidly evolving model performance demands increasingly frequent benchmark updates and manual curation processes have become unsustainable, highlighted by the accuracy surge on GSM8K (Grade School Math 8K) from 74\% to 95\% within two years.
Automated pipelines can address both the cost and efficiency challenges inherent in dataset creation. Another benefit of automation is its potential to mitigate data contamination, where test data are inadvertently exposed during pre-training or post-training, leading to overestimated performance. In response, automated approaches can be one of the solutions, which continually updates or refines test sets, known as dynamic benchmarks, ensuring that no test data are seen in advance. 
On the judgment side, as mentioned above, the shift to user prompts brings more open-ended responses, which pose further complexities: human judgment is expensive. 
Automated evaluators (i.e., ``LLMs-as-a-judge'') not only show promise in providing reliable, efficient assessments but also can produce more detailed, fine-grained evaluations of human-like responses. In Section~\ref{sec:autoeval}, we provide a comprehensive survey of these automated methods.

Although researchers have made significant progress along the two transitions outlined above, we argue that a fundamental contradiction persists between the training paradigm implied by scaling laws and the bounded evaluation practice. As model parameters, training FLOPs and data increase, the performance can be improved seemingly without bound. However, evaluation datasets cannot be expanded or diversified unbounded in practice considering the efficiency. That being said, current evaluation pipeline do not scale in tandem with model capabilities. The result is a growing mismatch between what models can do and what our tests can cover. This tension underlies many known challenges in LLM evaluation.
Take data contamination as an example, because the limited testing dataset can cover only a subset of a model’s capabilities, different models may gain heterogeneous advantages during evaluation, leading to unfair comparisons. That is, if a model has encountered and memorized the test samples during training, its measured abilities will align perfectly with what the dataset evaluates, granting it an outsized edge that does not necessarily reflect stronger true capabilities.

We designate the above problem --- how to leverage a bounded evaluation pipeline to assess an unbounded model capacity --- as the evaluation generalization issue. In other words, existing evaluation tend to concentrate on capabilities that models already exhibit or that can be expressed by a fixed test set, inherently limiting the scope. Thus, the core challenge of evaluation in the era of LLMs is to develop generalizable evaluation methods capable of anticipating future or unexpressed abilities.
In this survey, we examine this challenge from different perspectives: datasets, evaluators, and metrics, and explore potential solutions.
For example, some work focuses on predictive evaluation that carefully curates various tasks to estimate the performance of larger scale models based on that of smaller ones~\cite{srivastava2023beyond}. Or, Cao et. al.~\cite{cao2025revisiting} propose to combine performance and a new interpretability-based metric, Model Utilization Index (MUI), for evaluating the potential of LLMs beyond the given datasets. The basic idea mirrors human assessment practices: when judging an individual’s overall ability, we consider both the result and the effort required (i.e., MUI) --- less effort for equal performance denotes greater proficiency.

It is important to acknowledge that LLM evaluation is a rapidly evolving field. While we have endeavored to catalog the latest work on text‑centric evaluations, many studies remain at the preprint stage. Consequently, our emphasis here is on forward‑looking insights and research directions. Inevitably, some omissions or inaccuracies may occur.
We plan to maintain a dedicated GitHub repository and invite the community to help us for refinement; major contributors will be gratefully acknowledged or invited as collaborators.
\section{Core Capabilities and Datasets}
\label{sec:bench}
As LLMs unify a wide range of tasks, the first type transition is from task-specific to capability-based assessment.
In this section, we first discuss five core capabilities: knowledge, reasoning, instruction following, multimodal, and safety, with corresponding datasets, followed by their intersections and current live leaderboards. An illustration is shown in Figure~\ref{fig:data_tax}.
The Github page we will maintain and welcome any collaborators is \url{https://github.com/ALEX-nlp/Benchmark-of-core-capabilities/tree/main}.

\subsection{Knowledge Evaluation}

\begin{figure}
\center
\includegraphics[width=0.8\columnwidth]{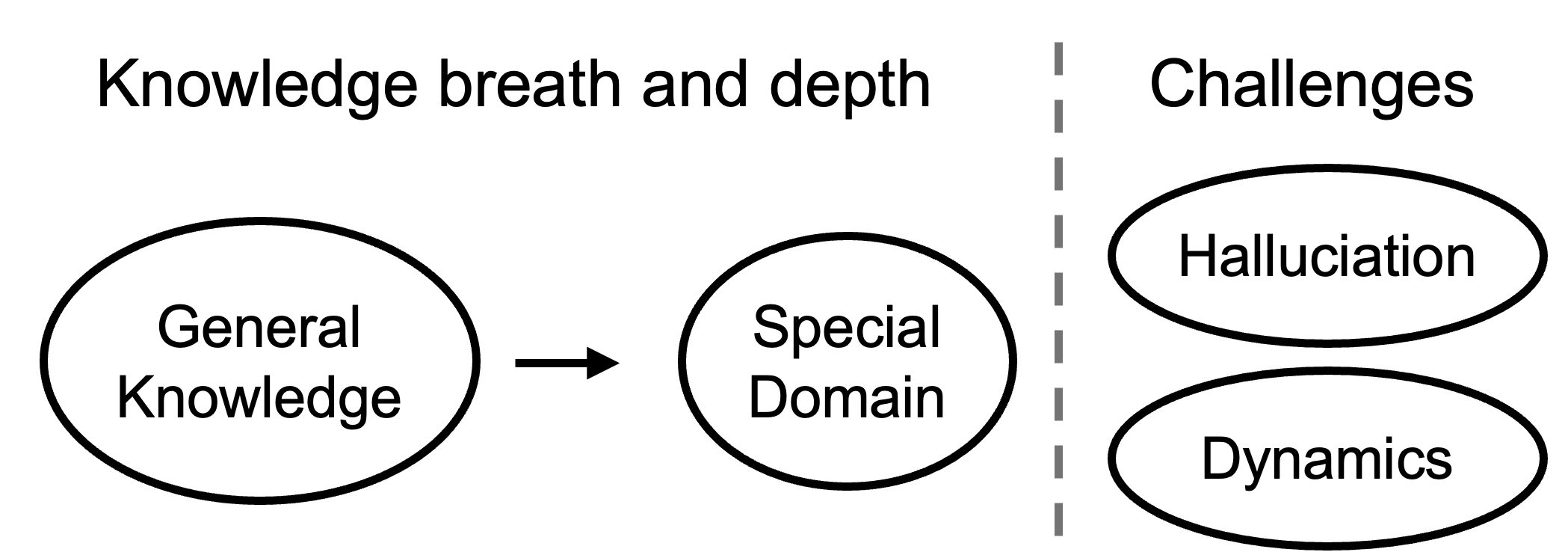}
\caption{
        The logic of reviewing knowledge evaluation.
        }
\label{fig:knowledge}
\end{figure}

Knowledge evaluation 
determines the models' ability to accurately recall, understand, and utilize factual information or human priors. Ensuring that LLMs possess a robust and reliable knowledge base is crucial for applications where precision and correctness are paramount. For example, it focuses on knowledge-intensive questions such as \textit{Who is the president of the United States?} or \textit{What is the capital of France?}.

In the early stage, benchmarks were primarily designed to assess the breadth of world knowledge in LLMs~\cite{petroni2019language}. These benchmarks focused on general knowledge derived from sources such as Wikipedia, ConceptNet, and other knowledge bases, typically adopting a question-answering format—sometimes accompanied by supporting documents. Representative examples include \textsc{TriviaQA}~\cite{joshi-etal-2017-triviaqa}, \textsc{NaturalQuestions}~\cite{kwiatkowski-etal-2019-natural}, \textsc{WebQuestions}~\cite{berant2013semantic}, and \textsc{CommonsenseQA}~\cite{talmor-etal-2019-commonsenseqa}. With the advent of LLM scaling laws, vast pre-training corpora have endowed these models with an extensive repository of general knowledge. Consequently, research attention has increasingly shifted from evaluating the breadth to probing the depth of domain-specific expertise in areas such as finance~\cite{islam2023financebench,xie2024finben,nie2024cfinbench} and law~\cite{lawbench,LexEval,li2024legalagentbench}.

Besides the breath and depth, knowledge evaluation also faces two major challenges. First, when it comes to unfamiliar or conflicting knowledge, LLMs may not admit their lack of understanding as humans do; instead, they may fabricate information, resulting in hallucinations. Worse still, as LLMs continue to evolve, they might even learn false information from the Internet. Consequently, LLMs could generate a large number of incorrect answers that are deceptive and potentially misleading to humans, which requires serious attention.
To address this issue, \textsc{TruthfulQA}~\cite{lin-etal-2022-truthfulqa} collected a set of well-known false claims or misconceptions, while \textsc{HaluEval}~\cite{li-etal-2023-halueval} curated questions that have no answers or are impossible to answer, requiring the model to point out that the question is unanswerable instead of generating a fake answer.

The second challenge concerns the dynamic nature of knowledge. Early datasets emphasized the timeliness and chronological order of knowledge~\cite{chen-etal-2021b-timeqa,zhang-choi-2021-situatedqa,liska-etal-2022-streamingqa,kasai-etal-2022-realtimeqa,yu2024kola,liu-etal-2024-untangle}, whereas later datasets focused more on addressing the data contamination issue using the latest knowledge from News articles~\cite{livebench}, Wikipedia~\cite{tang2024evowiki,wu2024antileak}.
The motivation is to accurately assess model advancements, evaluation datasets need to be continuously updated to prevent false negatives caused by outdated information.
Moreover, some scholars argue that consistently updating data can prevent performance overestimation due to data contamination, since as long as the evaluation data pertains to the latest knowledge, the model would not have been exposed to it, thereby eliminating data contamination issues. However, other scholars point out that the risk lies in the difficulty of completely distinguishing new from old knowledge based on a specific cutoff date (such as the model's release date). For instance, even if a movie is released after this date and the model should not have seen it, necessary information might have already been exposed to the model through early promotions and related activities~\cite{tang2024evowiki}. We will detail dynamic datasets in Section~\ref{sec:dybench}.

\subsection{Reasoning Evaluation}
Reasoning is a core component of intelligence in applying logic by drawing valid conclusions from new or existing information. Its evaluation is the key to gauge the true cognitive abilities of a model, such as problem solving, decision-making and human-like thought process. 
However, reasoning cannot be fully evaluated from a single perspective. Instead, researchers have developed methods to assess reasoning across multiple dimensions. In the following, we highlight several key domains: mathematics, coding, commonsense, long-context understanding, logic, planning, and miscellaneous tasks.

\subsubsection{Mathematics Evaluation}
\begin{figure}
\center
\includegraphics[width=0.5\textwidth]{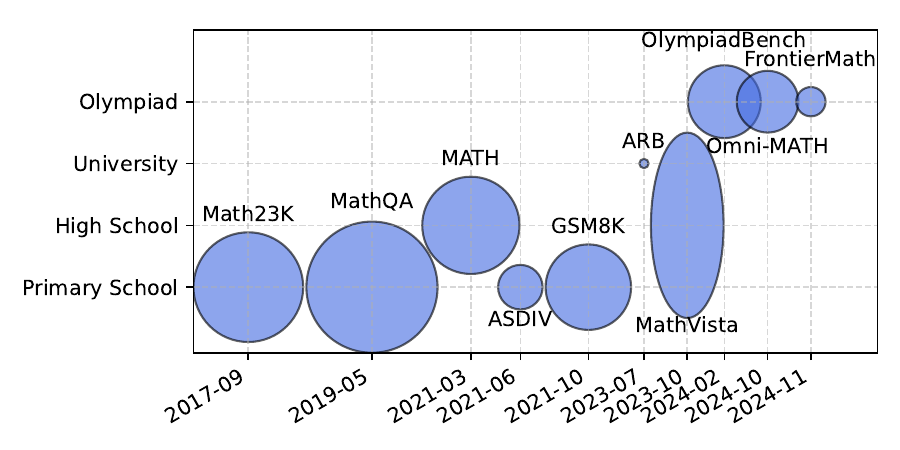}
\caption{Mathematics benchmarks in chronological order. The y-axis represents four difficulty levels: Primary School, High School, University, and Olympiad. The area represents the size of each dataset.}
\label{fig:math}
\end{figure}

Mathematical reasoning represents one of the most rigorously scrutinized aspects of reasoning evaluation. Its structured and precise reasoning process facilitates straightforward assessment. Furthermore, mathematics, as a cornerstone of abstract thought, is indispensable in scientific research, engineering, and related fields. As illustrated in Figure~\ref{fig:math}, mathematics benchmarks have evolved alongside the advancements in LLM capabilities, progressing from primary school-level problems to challenges of Olympiad-level difficulty.
In particular, before 2021, mathematics datasets mainly focused on primary school-level problems, reflecting the limited capabilities of language models at that time~\cite{wang-etal-2017-deep,amini-etal-2019-mathqa,miao-etal-2020-diverse}, where GSM8K~\cite{cobbe2021training} is still widely used.
After 2021, research efforts shifted toward high school~\cite{mathvista} or university-level problems~\cite{arb}. Example datasets include MATH~\cite{hendrycksmath2021}, which comprises 12,500 advanced high school math competition problems annotated with five difficulty levels.
Since 2024, the rapid advancement of LLMs has prompted researchers to further escalate the difficulty of benchmarks to the Olympiad contest level, aiming to extend the boundaries of these models~\cite{omnimath,olympiadbench}. 
Currently, the most difficult dataset is FrontierMath~\cite{frontiermath} crafted by expert mathematicians, covering major branches of modern mathematics --- from computational number theory to abstract algebraic geometry --- and often requiring hours or even days for specialists to solve. Even the most advanced reasoning LLMs like OpenAI o1 can only achieve around 3\% accuracy.

\subsubsection{Coding Evaluation}

\begin{figure}
\center
\includegraphics[width=\columnwidth]{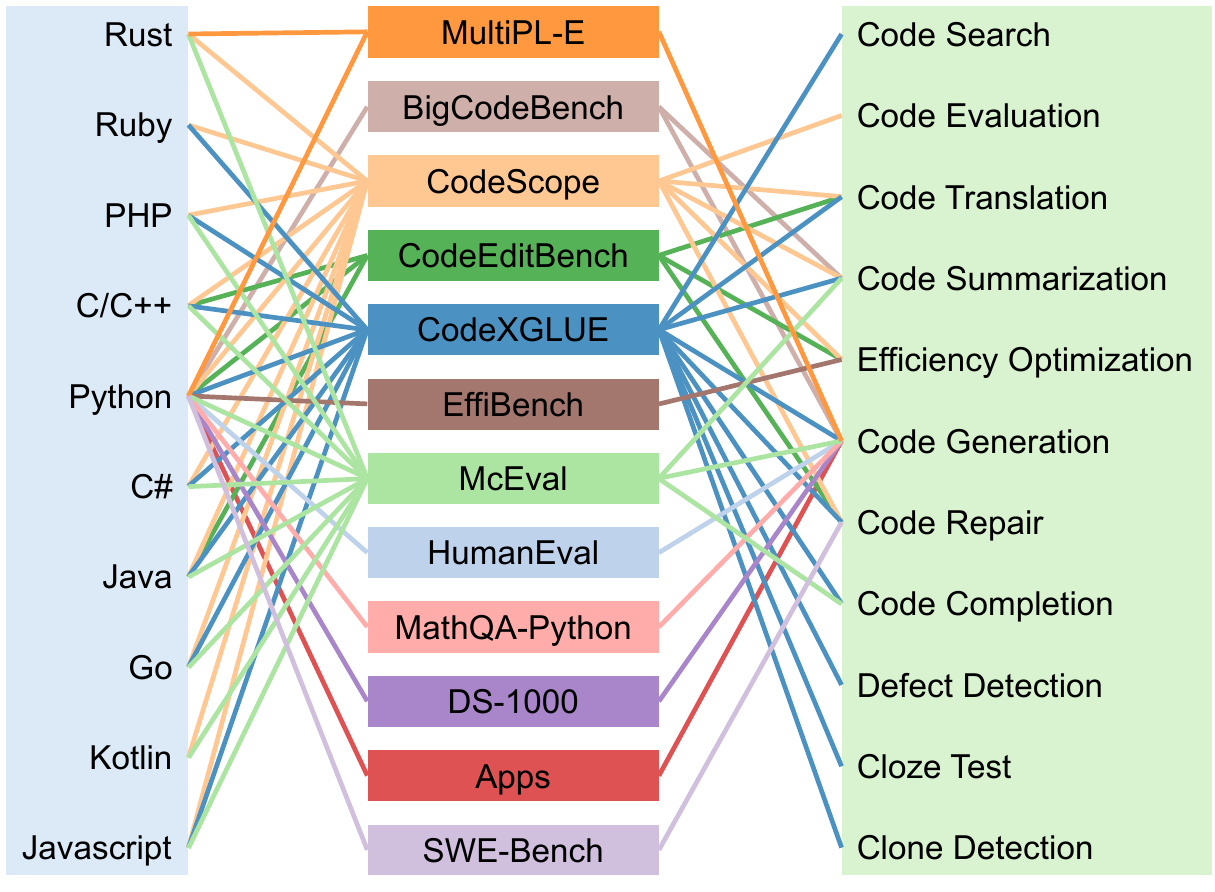}
\caption{Coding benchmarks, the programming languages (left), and coding tasks (right) that each benchmark use. We assign one color to each benchmark.}
\label{fig:code}
\end{figure}

Coding is another widely used reasoning evaluation task. Compared with mathematics, its reasoning process (i.e., code snippets) is also highly rigorous yet holds significant practical applications.
A variety of benchmarks have been introduced to evaluate LLMs' capabilities in code understanding and generation. We provide a high-level categorization of these benchmarks by programming languages and the primary coding tasks evaluated in Figure~\ref{fig:code}.
Observe that most benchmarks concentrate on Python and code generation task, given its wide adoption in both industry and academia~\cite{hendrycksapps2021,austin2021program,Lai2022DS1000,jimenez2024swebench,zhuo2024bigcodebench,effibench}.
To further evaluate cross-lingual capabilities, several benchmarks feature tasks in multiple programming languages~\cite{multiple,chen2021codex,codescope,mceval}.
Beyond language diversity, some benchmarks explicitly focus on different aspects of the software development life-cycle, including debugging, clone detection, defect detection, code completion, code-to-code translation, and requirement switching~\cite{DBLP:journals/corr/abs-2102-04664,codeeditorbench}.

Except for precise evaluation criteria, coding tasks also have strong practical values, which are becoming a high‑visibility benchmark for LLM reasoning. Nevertheless, they remain especially vulnerable to data contamination --- vast repositories of public code are ingested during pre‑training. To mitigate this, Livecodebench~\cite{livecodebench} continuously ingests newly released problems from coding competitions on platforms such as LeetCode, AtCoder, and Codeforces. By annotating each problem with its official release date, Livecodebench ensures that test items were unavailable during a model’s pre‑training period, effectively preventing contamination and overfitting and yielding a trustworthy, time‑aware assessment of coding performance.

\subsubsection{Logic Reasoning}
Another group of reasoning task is logic reasoning, usually involving three types: deduction (drawing conclusions), induction (recognizing patterns), and abductive reasoning (forming explanations). 
Logic reasoning is similar with math or coding in its well-defined nature, yet focuses on domain-independent inference patterns. This means that a model is expected to follow a structured reasoning process given the information at hand without any prior knowledge. Therefore, when benchmarking logic reasoning, except for real-world scenario, many attempts build a confining setting to minimize the unfair advantage of accumulated knowledge or learned information.

Deductive reasoning proceeds from general premises or rules to a guaranteed specific conclusion. If all premises are true and the logical steps are valid, the conclusion must be true, as in classical syllogisms or formal proofs. For example, given premises ``All birds can fly'' and ``Magpie is a bird'', a deductive model infers ``Magpie can fly''.
In the context of LLM evaluation, deductive benchmarks often involve determining whether a hypothesis holds true or false from provided premises~\cite{folio} or producing a step-by-step proof~\cite{ProofWriter}.
We can see that such tasks require the model to carry out multi-step logical derivations without introducing outside knowledge, while there are also other studies that curate probing benchmarks across different domains towards practical values, such as everyday situations~\cite{prontoqa} or exams~\cite{LogiQA, reclor}. 

Inductive reasoning is essentially the inverse of deduction: it draws general conclusions or rules from specific observations or instances. Here, the inference is probabilistic rather than certain --- the conclusion goes beyond the information provided. For example, given observations ``Magpie is a bird'' and ``Magpie can fly'', an inductive model may hypothesize ``All birds can fly'', which could later be proven wrong by a counterexample (``Ostrich cannot fly'').
Clear, the more the observations, the higher probability the inferred hypothesis holds true. This type of reasoning is easily influenced by prior knowledge. 
Thus, ARC-AGI benchmark~\cite{chollet2019measure} only assume core knowledge priors (``cognitive building blocks that are either present at birth or acquired very early in human development with minimal explicit instruction'') and design problems in a formal setting: given a set of input–output examples specifying some behavior for recognition, which is further simplified to 1-D pixel pattern in images~\cite{1darc}. Similarly, syntax-guided synthesis (SyGuS)~\cite{sygus} setup the task based on string transformation, and CLUTTR~\cite{CLUTRR} focuses on relational logic in narratives.

Abductive reasoning, also known as explanatory reasoning, involves generating the most plausible explanation for a given set of observations or facts.
Clearly, certainty is not guaranteed. The proposed explanation is a guess that could be wrong, but unlike induction, the goal of abductive reasoning is not a general rule but rather a specific hypothesis that accounts for the data. For example, given the observation ``The road is wet'', we may guess ``it probably rained recently''.
Such guessing heavily relies on the experience, so benchmarks for abductive reasoning often requires commonsense (which will be detailed next section) and an understanding of likely causal chains in everyday scenarios.
For example, $\alpha$NLI~\cite{acr} select the task of story completion and targets the more plausible connective explanation for how the characters got from the start to that end. To further challenge the deep abductive reasoning capabilities, True Detective benchmark~\cite{TrueDetective} setup the questions in murder-mystery narratives to ask who the crime is or what explains the mystery.
Considering the impacts of prior knowledge, there are also attempts like AbductionRules~\cite{AbductionRules}, which constructs synthetic logic puzzles for abduction. It presents a knowledge base of facts and rules (expressed in natural language) along with an ``unexpected'' observation, and the task is to hypothesize a missing fact or rule that would explain the observation.

Some recent surveys~\cite{xu2025largereasoningmodelssurvey,liu2025logical} and comprehensive benchmarks~\cite{bigbench} 
target all three types of reasoning. An interesting finding is that LLMs often perform the best for abductive reasoning and the worse for inductive reasoning.
But it is still challenging how to design benchmarks that truly measure reasoning and not just language proficiency or shallow pattern matching. 
As mentioned above, one proposal is to abstract away rich semantic content in tasks, so that an LLM’s performance reflects its grasp of reasoning structure rather than any prior knowledge. For example, using arbitrary symbols or ``neutral'' facts prevents the model from relying on memorized world knowledge, forcing it to rely on pure logic~\cite{logiglue}.
Indeed, models achieved high scores by learning the formal patterns, but such content-abstracted benchmarks have limits: they risk oversimplifying language understanding and may introduce unnatural regularities that models can exploit but that don’t translate to real-world reasoning~\cite{liu2025logical}.
On the other hand, benchmarks within some domains like science exams or detective stories ensure that models must deal with realistic language and background knowledge, but then it becomes harder to disentangle logical reasoning from domain knowledge.
The field is grappling with this balance between symbolic abstraction and natural complexity when evaluating reasoning.
Besides, as the True Detective~\cite{TrueDetective} results indicate, scaling reasoning to long contexts or more complex problems is still an open problem. Future benchmarks will likely need to push beyond toy tasks and short paragraphs, testing whether LLMs can maintain logical coherence over extended reasoning chains or in interactive, multi-turn settings.

\subsubsection{Commonsense Reasoning}
\label{sec:common_reason}

Commonsense reasoning refers to the fundamental level of practical knowledge and reasoning about everyday situations and events that is widely shared among people. Sometimes it adopts the same form of logic reasoning with commonsense knowledge.
Still, it is essential not only for humans to navigate daily life and interact with one another but also for artificial intelligence (AI) systems to better understand human needs and actions.
In terms of scenario, we can roughly categorize the evaluation into three domains: social, temporal, and physical commonsense. 
1) Social commonsense involves understanding interpersonal interactions and human behavior. Representative datasets in this category include Naive Psychology~\cite{naive}, ROCStories~\cite{rocstory}, Social IQa, the Winograd Schema Challenge (WSC)~\cite{sakaguchi2021winogrande}, Choice of Plausible Alternatives (COPA)~\cite{COPA}, VCR (visual commonsense reasoning)~\cite{VCR}, and e-CARE~\cite{ecare} (explainable commonsense).
2) Temporal commonsense pertains to the sequencing of events, causality, and time-related inferences like duration, frequency, or ordering. Key datasets here include MCTaco~\cite{ZKNR19}, UDS-T~\cite{udst}, and MavenERE~\cite{wang-etal-2022-maven}. 
3) Finally, physical commonsense encompasses fundamental knowledge about the physical world, including object properties and spatial relationships, such as Physical IQa~\cite{PIQA}, HellaSwag~\cite{zellers2019hellaswag}, Abductive NLI~\cite{acr}, SWAG~\cite{SWAG}, CommonsenseQA~\cite{talmor-etal-2019-commonsenseqa}, and JHU Ordinal Commonsense~\cite{Ordinal}.

Besides datasets, there are also many commonsense resources available for both training and evaluation purposes. Early projects were primarily developed by human experts including Cyc~\cite{cyc} and OpenCyc~\cite{opencyc}. These systems encoded ontological relationships between objects using formal logic, categorizing entities into types such as entities, sets, functions, and truth functions, and contained a wealth of commonsense assertions.
Concurrently, a team at the MIT Media Lab developed the Open Mind Common Sense project~\cite{omcs}, later evolving into ConceptNet~\cite{conceptnet}. This project harvested online data and integrated diverse knowledge sources. The latest version, ConceptNet 5.5, employs automated extraction techniques and comprises over eight million nodes and more than twenty-one million links, incorporating multilingual resources as well as connections to other knowledge graphs.
Among those automatically curated resources~\cite{aser,mmekg}, a notable one is ATOMIC~\cite{atomic}, a crowd-sourced knowledge graph that features textual descriptions for around 300,000 event nodes and approximately 877,000 ``if-event-then'' triplets, capturing nine distinct types of causal relationships between everyday events. Building upon these foundations, the VisualComet project~\cite{visualcomet} extended the realm of commonsense reasoning into the multimodal domain by proposing Visual Commonsense Graphs.
Additionally, there are domain-specific or purpose-oriented commonsense reasoning resources, such as for sentiment analysis~\cite{SenticNet}, causal reasoning~\cite{CauseNet}, and e-commerce intention~\cite{FolkScope}.

Overall, commonsense reasoning has a long research heritage, supported by extensive resources and benchmark datasets. More recently, the emergence of LLMs exhibiting super-human performance on certain commonsense tasks. However, this does not imply that the challenge has been completely resolved. In real-world applications, LLMs still lag behind human capabilities when it comes to complex commonsense reasoning, especially in multimodal tasks.
Two core challenges underlie this gap:
1) commonsense knowledge is never explicitly stated in text, images or other modalities, which hampers acquirement or robust reasoning;
2) Unlike mathematics or coding, commonsense tasks do not possess clear formal structure or single ``correct'' answers. This ambiguity complicates both the construction of datasets and the accurate evaluation of open‑ended responses.
To conclude, commonsense task may be the key in extending reasoning evaluation from clear, well‑formed tasks to more ambiguous, open-world problems in the near future.

\subsubsection{Long Context Reasoning}~\label{sec:lcr} 
Text remains the primary medium for interacting with LLMs, and knowledge can be embedded in long texts in diverse forms. As a result, the ability to reason over extended contexts has become a critical capability. Long-context reasoning refers to increasing the input length that LLMs can process, while ensuring they can effectively understand, learn from, and reason over the information contained in these longer inputs.

We classify it as reasoning because, compared to methods like retrieval-augmented generation (RAG), which are more suited for extractive or localized tasks, long-context reasoning excels in scenarios where the model needs to perform global reasoning by leveraging all the input information.
Also, this is highly significant in practical applications, as it serves both as an important means for integrating external knowledge, recording historical behaviors or interactions, and following complex instructions~\cite{ragnlp}.

To propel this line of work, long‑context LLMs have adopted techniques such as interpolation~\cite{interpolation}, extrapolation, fine‑tuning, and architectural optimizations to rapidly extend their supported token windows. Correspondingly, evaluation benchmarks continuously raise both the maximum input sequence length and the complexity of tasks, to ensure that assessment keeps pace with ever‑growing model capabilities.
At first, L-Eval~\cite{an2023leval} and LongBench~\cite{bai2024longbench} are at a moderate scale --- contexts from roughly 3K to 60K tokens, including tasks like single-document QA, multi-document QA, summarization, and code completion.
Then, InfinityBench~\cite{zhang-etal-2024-bench} steps up to ultra‑long contexts (about 
100K tokens) at domains such as novel and coding.
Recently, LongBench v2~\cite{longbenchv2} is at the extreme frontier, which pushes to 2 million words of context across 503 questions in six categories: single‑ and multi‑document QA, extended in‑context learning, long‑dialogue comprehension, large code‑repository understanding, and structured‑data reasoning. This benchmark emphasizes deep logical inference, cross‑document linking, and structured‑data extraction at unprecedented scale.

Although there are many efforts mentioned above, the definition of ``context" is still not clear enough~\cite{li2024longcontextvsrag}, which dictates both dataset structure and evaluation focus when designing long‑context benchmarks. If context is 
in the form of a single coherent document, such as a novel or research paper, the benchmark must ensure tight question–passage alignment and test a model’s ability to integrate clues spread across multiple sections. These datasets probe deep reading comprehension and multi‑paragraph reasoning but are costly to curate, subject to copyright constraints, and slow to refresh. Conversely, if context is 
provided as an artificial concatenation of shorter excerpts, e.g., simply grouping Wikipedia articles together, benchmarks can be scaled quickly. Yet the relevance between questions and information becomes uneven, and the task shifts toward retrieving salient facts amid noise, exposing failures such as position bias or the ``lost‑in‑the‑middle'' effect.

This dual interpretation poses three core challenges for evaluation. First, benchmarks must strike a balance between reading and retrieval skills, ensuring that neither devolves into trivial keyword matching nor pure long‑span memorization. Second, they require reliable metrics of question–context relevance; without such controls, high scores may reflect chance matches rather than genuine understanding. Third, as token windows expand, benchmarks must evolve dynamically: coherent long texts are hard to source continuously, while concatenated corpora risk overlapping with pre‑training data, demanding fine‑grained de‑duplication and release‑date tagging to prevent contamination. Addressing these challenges is essential for keeping long‑context evaluation both realistic and forward‑looking. On the other hand, while long-context reasoning can be partially reflected in benchmark results, the specific reasoning capabilities assessed may be domain-specific, depending on the nature of the documents used in the benchmarks.

\subsubsection{Planning}~\label{sec:data_plan}
Planning is a special type of reasoning. Instead of inferring new knowledge from existing ones, it aims at decompose high-level objectives to fine-grained, relatively simple steps.
Due to the task's complexity, this process usually needs to combine various reasoning skills, setting a high bar for the model. Nevertheless, planning is the key for models dealing with dynamic and complex tasks then stepping from simple textual contexts, to virtual environments and to physical worlds. According to the model's working environments, we examine three dimensions of planning benchmarks: (1) \textit{task planning} for textual goal decomposition, (2) \textit{agent planning} for autonomous decision-making in virtual, interactive environments, and (3) \textit{embodied agent planning} that integrates physical interaction with spatial reasoning.

Textual task planning focuses on generating structured sequences of steps to achieve specified goals, often requiring hierarchical decomposition. Early work in Goal-Oriented Script Construction (GOSC)~\cite{gosc} established baselines using the WikiHow~\cite{wikihow} benchmark for step sequence generation, later extended by Instructables~\cite{hsg} to incorporate hierarchical subgoals. Subsequent benchmarks like PlanBench~\cite{PlanBench} systematically evaluate validity and cost-optimality of generated plans, revealing significant gaps between LLM capabilities and human reasoning. TaskBench~\cite{taskbench} introduces tool invocation graphs to assess execution consistency alongside planning precision. Real-world applications are explored through Natural Plan~\cite{NATURAL} for trip scheduling and meeting coordination, WorkBench~\cite{workbench} for digital workplace task management, and UltraTool~\cite{ultratool} for end-to-end tool utilization in complex problem-solving scenarios.

Virtual agent planning evaluates autonomous systems' capacity to navigate in simulated or digital environments. WebShop~\cite{WebShop} targets e-commerce simulations requiring complex query interpretation and purchase optimization across 1.18 million real products.
TravelPlanner~\cite{TravelPlanner} chooses requiring agents to balance budget, logistics, and commonsense constraints while coordinating multiple information sources for travel itineraries.
SmartPlay~\cite{smartplay} tests adaptive reasoning through six game environments requiring spatial and strategic planning. 
Theoretical foundations are strengthened by the tri-modal evaluation framework (autonomous/heuristic/human-in-the-loop) and the benchmark for action reasoning and plan reuse~\cite{ontheplan,cantplan}. Robotic integration is pioneered by SayCan~\cite{saycan}, which grounds planning in physical affordances for real-world mobile manipulation.

Embodied agent planning bridges digital reasoning with physical world, demanding tight integration of perception, environment interaction, and actions. According to the key components, we discuss the evaluation from two perspectives of environment and planning types.
In terms of environments, embodied planning benchmarks span from abstract symbolic worlds to high-fidelity 3D simulations.
For symbolic or text-based virtual environments, which is different from the aforementioned task planning and virtual agent planning, they leverage descriptive language or abstract state representations for rooms and objects instead of seeing pixels~\cite{textworld}.
For 3D simulations, many benchmarks adopt first-person view based on simulators like Habitat~\cite{Habitat} and iGibson~\cite{li2021igibson} and an agent must interact with objects or navigate spaces. This type of evaluation emphasizes photorealism and physics, featuring realistic lighting, textures, and physical object dynamics. 
Examples include ALFRED~\cite{alfred} and BEHAVIOR~\cite{BEHAVIOR-1K}.
By combining both, ALFWorld~\cite{alfworld} is a hybrid platform that aligns a text-based world with the 3D tasks from ALFRED, allowing agents to practice in a simplified symbolic setting before transferring to a realistic simulator.

In terms of planning types, different benchmarks demand different levels of planning granularity. Low-level action planning requires sequences of fine-grained actions (navigation steps, motor primitives). For example, an agent needs to plan a path through a 3D scene, issuing low-level motions (forward, turn) to reach a target coordinate or object~\cite{Habitat}. This is often framed as visual navigation and tests short-term planning and obstacle avoidance, albeit potentially over long distances. 
In contrast, high-level task planning involves deciding on a sequence of sub-tasks or goals to satisfy an overall objective, e.g., ``clean the coffee cup and put it back''. Examples datasets include ALFRED~\cite{alfred} and BEHAVIOR~\cite{BEHAVIOR-1K}.
Although high-level tasks have achieved promising results, some studies argue the potential overestimation and delve into single-step planning~\cite{mu2021maniskill,wang2024navigating}. Their analysis reveals both types of evaluation are important since notable drawbacks like numerical comprehension, heavy selective biases over directional concepts, or recurrent issues, still exist and may be critical when transferring to real world.

\subsubsection{Miscellaneous Reasoning}
Apart from the aforementioned reasoning tasks, there exists a diverse range of reasoning skills that we collectively refer to as miscellaneous reasoning. Example symbolic reasoning tasks include coin flip reasoning and last letter concatenation~\cite{wei2022chain}. The basic idea is to define a set of transformation rules, and the model is required to apply these rules systematically to infer the correct outcome given an initial state. Similarly, there are also some common IQ test puzzles and algorithmic problems, such as classic puzzles like the tiger-eats-sheep problem or the gold division problem. Clearly, these tasks are in-between logical reasoning or instruction-following.

Other examples include visual reasoning and designs  tasks defined in~\cite{rahmanzadehgervi2024vision}. These tasks include counting line intersections, determining the relationship between two circles, identifying the circled letter, counting overlapping shapes, counting nested squares, counting the rows and columns of a grid, and following single-colored paths. Though simple, most vision language models (VLMs) perform unsatisfactory.

Spatial reasoning also attracts many research attention. The work~\cite{spatial} defines 2D and 3D trajectory labeling and relationship identification. This work employs the CALVIN benchmark~\cite{calvin}, which assesses LLMs in long-horizon, language-conditioned robotic manipulation tasks.

\subsection{Instruction Following}

\begin{figure}
    \centering
    \subfloat[Multi-task generalization.]{%
        \includegraphics[width=0.47\textwidth]{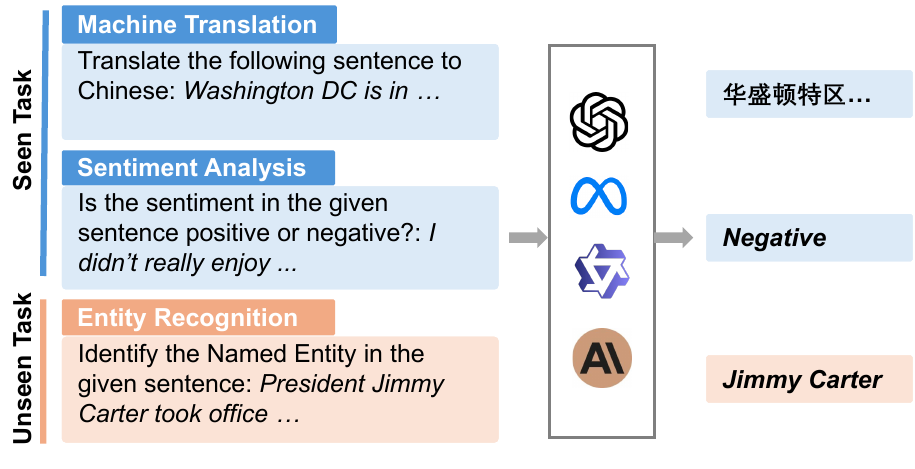}
        \label{fig:multi-task}
    }
    \hfill 
    \subfloat[Prompt generalization.]{%
        \includegraphics[width=0.47\textwidth]{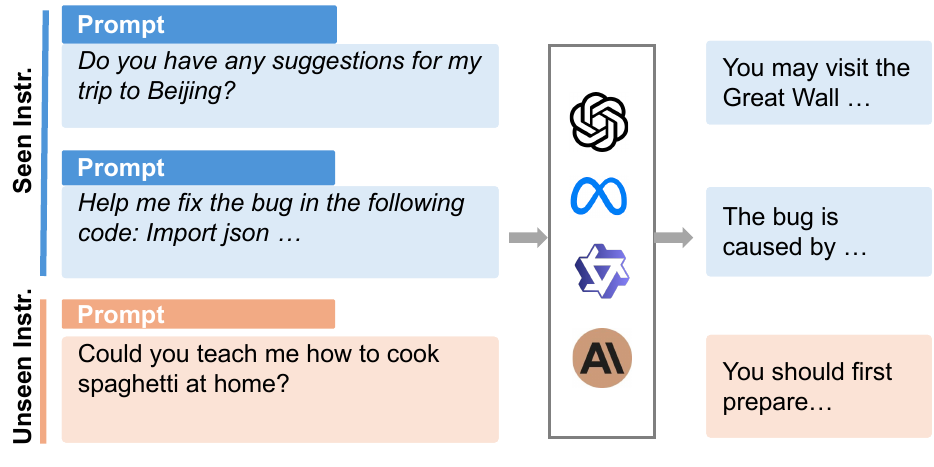}
        \label{fig:prompt-gen}
    }
\caption{
        Illustration of instruction following's paradigm shift: from tasks, where tasks are gathered and their descriptions are used as instructions, to user needs where user prompts are regarded as ``fine-grained" tasks' instructions.
    }
\label{fig:instruction-following}
\end{figure}

Instruction following aims to assess whether models can comprehend human inputs and provide appropriate responses. As model capabilities have advanced, this evaluation method has progressively evolved from conventional NLP tasks to diverse human needs (Figure~\ref{fig:instruction-following}).

In the early stages, models were limited to performing single tasks, with training focused on mapping inputs to outputs on specific datasets. To enhance generalization, multi-task learning was introduced, shifting model predictions from $p(y|x)$ to $p(y|x, task\ description)$~\cite{radford2019language}. Consequently, researchers began aggregating diverse tasks and crafting detailed task descriptions, resulting in multi-task fine-tuning datasets collectively known as instruction tuning. Held-out tasks not included in the training set were then used for evaluation. Early instruction following evaluations thus centered on traditional NLP problems, such as question answering, question generation, and text classification. Representative benchmarks in this phase included \textsc{Natural Instructions}~\cite{mishra-etal-2022-cross}, \textsc{TO-Eval}~\cite{sanh2022multitask}, \textsc{InstructEval}~\cite{chia-etal-2024-instructeval}, \textsc{Flan}~\cite{weifinetuned}
and \textsc{SuperNI-Test}~\cite{wang-etal-2022-super}.

As more tasks are included to enhance generalization, the goal of instruction tuning shifts to real-world human needs, from which modern NLP tasks are derived. Thus, task descriptions can be viewed as scientific definitions of those needs. Researchers then began collecting a diverse array of real-world user prompts, moving beyond the confines of specific tasks, such as \textsc{ShareGPT}~\cite{eccleston2023sharegpt}, \textsc{FreeDolly}~\cite{conover2023free}, \textsc{OpenAssistant}~\cite{openassistant}. 
The release of ChatGPT further accelerated this trend, as LLMs were integrated into online service environments that allow users to issue a wide variety of instructions framed as open-world ``tasks''. A representative example is \textsc{Chatbot Arena}~\cite{zheng2023judging}, renowned for its Elo rating score system. In this platform, users provide instructions to chatbots, and two randomly selected LLMs generate responses for direct comparison. Human annotators then select the better response in real time, with each outcome contributing to an evolving Elo rating that dynamically quantifies the relative performance of the models—thereby more closely mirroring real-world usage scenarios.

This shift from task-specific evaluations to user-driven needs ensures that models can handle a wide variety of instructions and respond effectively to the multifaceted demands encountered in practical applications. However, it also introduces significant evaluation challenges: responses become more open-ended and non-structured, making it difficult to achieve reliable and consistent scores through human evaluation, which is both resource-intensive and time-consuming. Consequently, researchers have begun exploring automated evaluation methods to improve efficiency and scalability.
For example, \textsc{AlpacaEval}~\cite{alpaca_eval} and \textsc{VicunaEval}~\cite{vicuna2023} have experimented with using LLMs to score the quality of responses. These studies found that LLMs can not only generate relatively consistent scores but also provide detailed explanations for those scores. Furthermore, benchmarks such as \textsc{Arena Hard Auto}~\cite{li2024crowdsourced} have been developed. In these benchmarks, user instructions collected from online environments are evaluated in a pairwise manner --- similar to \textsc{Chatbot Arena} --- where two LLMs generate responses to the same instruction. The key difference is that evaluations are conducted by a powerful LLM, such as GPT-4, rather than human annotators, thereby improving scalability and efficiency~\cite{zheng2023judging}.

Nevertheless, this automated approach has faced criticism for introducing potential biases inherent in the judging LLMs. Such biases, including preferences for verbosity or specific response styles, may lead to unfair evaluation outcomes~\cite{li2024stylecontrol, chen-etal-2024-humans, dubois2024length, park-etal-2024-disentangling}. To mitigate these issues, style-control variants such as \textsc{StyleControl Arena}\cite{li2024stylecontrol} and \textsc{Length-Controlled AlpacaEval}\cite{dubois2024length} have been introduced. These benchmarks seek to disentangle stylistic factors from the substantive content of responses, enabling fairer comparisons between LLM outputs. A more detailed discussion of automated evaluation methods will be presented in the Section~\ref{sec:autoeval}.

Although pairwise comparison benchmarks are valuable for assessing relative performance, they do not provide fine-grained scores for specific capabilities. To address this limitation, a new class of benchmarks has been developed to evaluate instruction-following ability in an absolute and fine-grained manner~\cite{zhou2023instruction, jiang2023followbench, lin2024wildbench}. For instance, \textsc{IfEval}~\cite{zhou2023instruction} introduced 25 rule-verifiable constraints (e.g., Output your response in all uppercase letters, within 10 words, without using the word "I", etc.), requiring LLMs to generate responses that satisfy these constraints. \textsc{FollowBench}~\cite{jiang2023followbench} extended this idea by expanding the constraints from rule-verifiable to model-verifiable, wherein another strong LLM is tasked with verifying whether the generated responses meet the constraints. \textsc{WildBench}~\cite{lin2024wildbench} further advanced this approach by providing a human-annotated checklist for each instruction, with a judge LLM verifying whether the generated response satisfies the checklist --- thus adding an additional layer of human oversight.

\subsection{Multi-modal Evaluation}
Multimodal evaluation measures the ability of LLMs to process different data modalities beyond text, such as audio or tabular data. This capability enhances the versatility of AI models in real-world applications. Several survey papers~\citep{fu2023mme,li2024survey,huang2024survey} have focused on evaluations in this domain; therefore, we select visual information as a complementary aspect to the text-based capability assessments discussed above. Below are some representative tasks.

\subsubsection{Visual Question Answering} Visual Question Answering (VQA) is to answer questions based on both textual and visual information.
We start with basic visual perception tasks, like RealWorldQA~\cite{grok15} that evaluate real-world spatial understanding including counting, identifying, and locating objects in images. These tasks are easy for humans but still challenging for models.
To further assess cognitive abilities, \textsc{MME}~\cite{fu2023mme} designs reasoning, coding, and planning tasks.
\textsc{MMT-Bench}~\cite{ying2024mmt} dives deeper by decomposing visual abilities into 32 meta-abilities (e.g., counting, locating, and identifying) and constructing a comprehensive benchmarks.
Except for perception and cognition, MMMU~\cite{yue2024mmmupro} and MMMU-Pro~\cite{yue2024mmmupro} curated massive multi-discipline tasks demanding college-level subject knowledge. 
While, recent works have shown that many MMMU samples could be answered without visual information, this raises concerns about uni-modal bias~\cite{chen2024quantifying}. 
To address this issue, \textsc{MMStar}~\cite{chen2024we}, a vision-indispensable multi-modal benchmark, was proposed.
Each sample in \textsc{MMstar} is verified by human to ensure the visual content is essential to answer the question.
Furthermore, hallucination and long-tail issue are also considered in \textsc{MMBench}~\cite{MMBench} and HallusionBench~\cite{guan2023hallusionbench}, respectively.

\subsubsection{Visual Document Comprehension} Visual Document Comprehension regards document, including text, tables, and diagrams, etc., as visual inputs (e.g., images) for the following tasks. Compared with textual document understanding, this type of methods enjoy an efficient end-to-end manner and can achieve maximum retention of information. For example, there is no additional step to parse text from images and the layout is maintained.
Therefore, benchmarks are curated to evaluate text understanding from document screenshot~\cite{docvqa} or taken in the wild~\cite{textvqa,tang2024mtvqa}, infographic text comprehension~\cite{docvqa}, multi-modal document understanding~\cite{masry2022chartqa}, and scientific diagram comprehension~\cite{kembhavi2016diagram}.

\subsubsection{Multi-image Understanding} While earlier MLLMs are mostly trained to align single images with natural language components, one emergent capability that recent efforts seek to extend is multi-image understanding, or more broadly, interleaved processing of (multiple) images and texts.
In this context, earlier benchmarks including multi-image examples typically focus on specific scopes of reasoning and do not provide a comprehensive assessment~\cite{fu2024blink,fu2024blink,wang2024mementos,ying2024mmt,yue2023mmmu}.
Recent efforts assess MLLMs in multi-image scenarios. For example, MANTIS-Eval~\cite{jiang2024mantis}  is a human-annotated benchmark comprising 207 examples for multi-image reasoning, such as size perceptions and weight comparisons, while DEMON~\cite{li2024fine} evaluates whether MLLMs can follow zero-shot demonstrative instructions.
A milestone benchmark for this challenge is MuirBench~\cite{wang2025muirbench}. This comprehensive benchmark contains 11,264 images and 2,600 multiple-choice questions, evaluates on a range of 12 multi-image understanding abilities (e.g. geographic understanding, diagram understanding, visual retrieval, etc.) and 10 diverse multi-image relations (e.g. narrative, complementary, etc.).

Similarly, video understanding can be regarded as an extension of image understanding to a sequence of images, considering the temporal and spatial features among images. For example, MVBench~\cite{li2024mvbench} and PerceptionTest~\cite{patraucean2024perception} evaluates general video comprehension. Clearly, along with the increasing video length, MLLMs is required to processing massive images within the context window like ``visual long context reasoning'' (visual version of Section~\ref{sec:lcr}).
EgoSchema~\cite{mangalam2023egoschema} and Video-MME~\cite{fu2024video} target the comprehension of long-term video up to one hour.

\subsection{Safety}
Along with the increasing capabilities of LLMs, their deployment raises serious safety concerns. 
Safety evaluation aims at assessing a model’s ability to avoid generating harmful, unethical, or biased outputs, ensuring its alignment with human values and societal norms.
A recent survey~\cite{ma2025safety} classified existing works into various attack and defense groups, including adversarial attacks/defenses, backdoor \& poisoning attacks/defenses, jailbreak attacks/defenses, intellectual property protection, membership inference attacks, data extraction attacks, prompt injection attacks, etc.
While, another survey~\cite{safetyprompts} comprehensively introduce the open datasets and categorizes them into five main purposes: broad safety, narrowly defined safety, value integrity, bias, and other. 
Differently, our categorization is driven by analyzing evaluation trends and contains four directions: 1) content safety, 2) multi-dimensional trustworthiness, 3) adversarial robustness, and 4) agentic safety.

\subsubsection{Content Safety}~\label{sec:con_safe}
At the most fundamental level, content‑safety benchmarks probe whether an LLM can identify, refuse, or filter toxic, hateful, violent, or otherwise disallowed text under non‑adversarial conditions. Evaluations appear in three formats. The first one adopts single‑sentence classification, e.g., ToxiGen~\cite{hartvigsen2022toxigen} includes 274,000 machine-generated statements targeting 13 minority groups, each labeled as either toxic or benign.
Second, recent studies, like RealToxicityPrompts~\cite{gehman-etal-2020-realtoxicityprompts}, ToxicChat~\cite{lin-etal-2023-toxicchat}, BeaverTails~\cite{ji2024beavertails}, and DiaSafety~\cite{sun-etal-2022-safety}, mimic the settings in real-world applications, which collects prompt‑response pairs. Thus, evaluation can either treat it as text classification, the same as the first format above, or, third, feed the testing prompt into LLMs and leverage external tools to judge the newly generated response.
Clearly, two major challenges lie in the design of prompts and the performance of judge tools, which are the main focus of adversarial robustness as discussed later.
In addition, there is a growing emphasis on multilingual content moderation or specific domains such as gender and sexuality~\cite{fleisig-etal-2023-fairprism}.
To conclude, the key challenges for content safety benchmarks are two-fold: 
1) the hate speech may be nuanced and contain no obvious slurs or profanity, which motivates ToxiGen~\cite{hartvigsen2022toxigen} to design an adversarial classifier-in-the-loop generation process.
2) the hate speech should, as much as possible, originate from or closely resemble everyday life. Example datasets including RealToxicityPrompts~\cite{gehman-etal-2020-realtoxicityprompts} and DiaSafety~\cite{sun-etal-2022-safety} then collect data from real world like Reddit.

\subsubsection{Multi‑Dimensional Trustworthiness}
LLM ``safety'' is not a single metric.
Complementary to content toxicity or hatefulness as mentioned above, several recent benchmarks aim to evaluate LLMs holistically across multiple dimensions like bias~\cite{ganguli2023challenges}.
DecodingTrust~\cite{DecodingTrust} assembles tests for eight different aspects including toxicity, stereotype bias, privacy, ethics, fairness, as well as adversarial and out-of-distribution robustness.
HELM Safety\footnote{\url{https://crfm.stanford.edu/2024/11/08/helm-safety.html}} combines five benchmarks, covering six harm domains: violence, sexual content, harassment, self-harm, deception, and discrimination, and draws on specialized sub-benchmarks for each.
The AegisSafety dataset~\cite{aegis} define a broad taxonomy of 13 critical risk and 9 sparse risk categories.
SorryBench~\cite{SORRY-Bench} spans 45 fine-grained safety categories targeting refusal behaviors. Meanwhile, it includes multi-lingual variations, which is also highlighted by XSafety~\cite{xsafety} and S-Eval~\cite{S-Eval}.

In terms of evaluation format, most benchmarks follows similar settings with those for content safety and adopt multi-choice questions, such as SafetyBench~\cite{safetybench}, DecodingTrust~\cite{DecodingTrust}, SGBench~\cite{SG-Bench}.
To improve the difficulty levels, SALAD-Bench~\cite{SALAD-Bench} introduces attack and defense methods to enhance the prompts categorized into 6 domains, 16 tasks, and 66 specific categories.
While, CHiSafetyBench~\cite{CHiSafetyBench} designs a hierarchical benchmark across 5 risk areas and 31 categories to better organize the multiple safety dimensions.
Except for structured tests, scenario-based tests are gaining popularity for practical values. The model is placed in a concrete situation and must take a stance or choose an action consistent with safety or ethics. For instance, the HHH benchmark~\cite{bai2022training} compares pairs of model outputs in different interaction scenarios and asks which response better aligns with ethical values: Helpfulness, Honesty, and Harmlessness. This format, using human preference judgments on model outputs, checks if the model can be simultaneously useful, truthful, and non-harmful.
Another example is the ETHICS~\cite{hendrycksaligning}, which poses ethical dilemmas or scenario-based questions covering dimensions like justice, deontology, virtue ethics.
For better understanding model's safety, DoNotAnswer~\cite{wang2024not} provides an explanation for why a response should be refused, enabling evaluators to check not just if the model refuses, but whether it understood the risk. 

A core challenge in multi-dimensional trust evaluation is coverage and scalability. While, curating tests for every potential risk is labor-intensive. This has led to efforts to crowdsource\footnote{\url{https://github.com/openai/evals}} and automate scenario generation~\ref{sec:autoeval}. However, using LLMs to judge other LLMs can introduce error if not carefully validated~\cite{koh2024can}, which will be further discussed in Section~\ref{sec:multieval}.
Therefore, a clear trend in evaluation design for trustworthiness is moving beyond static question sets toward more interactive simulations. For example, some studies introduce role-playing games to simulate some scenarios, so that the involved agents may discover potential risks and produce training/testing data automatically~\cite{pang2024self}.

\begin{figure*}
\centering
\includegraphics[width=0.85\textwidth]{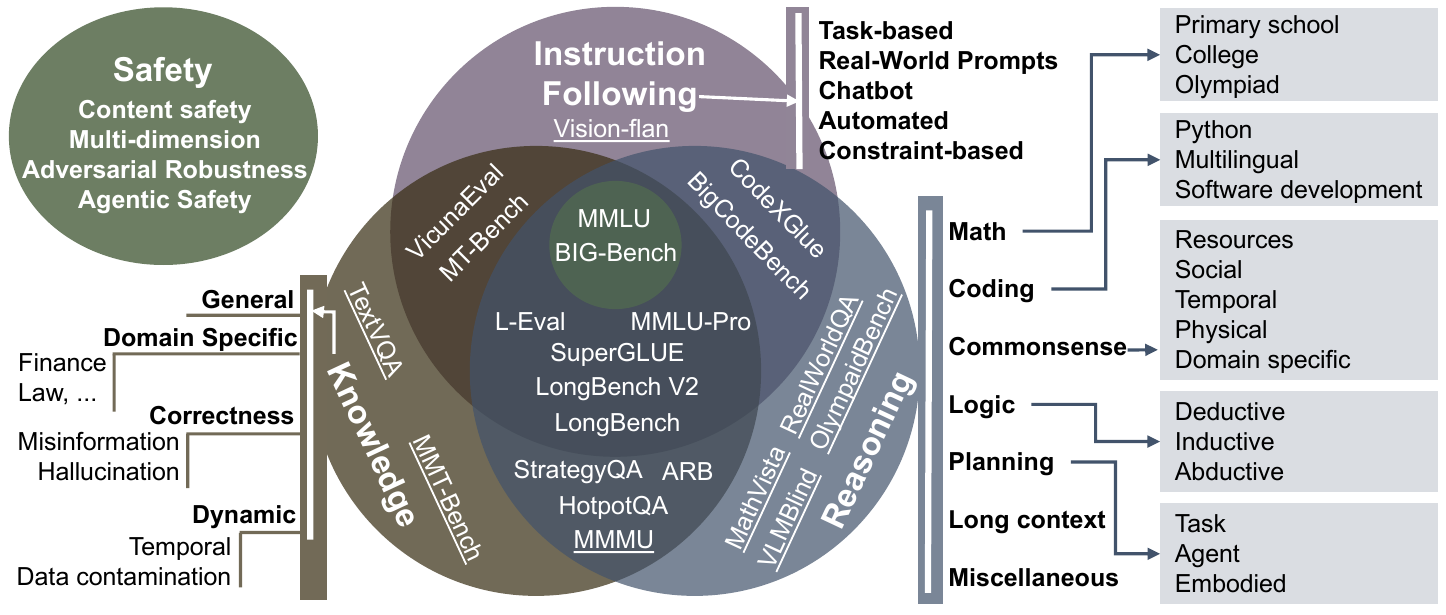}
\caption{Illustration of capability-based benchmark taxonomy involving: knowledge, reasoning, instruction following, multimodal, and safety. The interaction denotes the interplay of different abilities. Note that multi-modal benchmarks are marked with underline. The taxonomy and corresponding datasets are tentative and we will keep improve it.}
\label{fig:data_tax}
\end{figure*}

\begin{table*}[ht]
  \centering
  \caption{Capability‐based benchmark taxonomy.}
  \label{tab:cap-taxonomy}
  \renewcommand{\arraystretch}{1.2}
  \begin{tabular}{@{}p{0.32\textwidth}|p{0.32\textwidth}|p{0.32\textwidth}@{}}
    \toprule
    {%
      \begin{tabular}[t]{@{}r@{\quad}l@{\quad}r@{}}
        \multirow{11}{*}{\rotatebox{90}{\small\textbf{Knowledge}}}
          & \textbf{General}             & \cite{petroni2019language,joshi-etal-2017-triviaqa,kwiatkowski-etal-2019-natural,berant2013semantic,talmor-etal-2019-commonsenseqa} \\
          & \textbf{Domain:}             & \\
          & \quad Finance                & \cite{islam2023financebench,xie2024finben,nie2024cfinbench} \\
          & \quad Law                    & \cite{lawbench,LexEval,li2024legalagentbench} \\
          & \textbf{Incorrect:}          & \\
          & \quad Misinformation         & \cite{lin-etal-2022-truthfulqa} \\
          & \quad Hallucination          & \cite{li-etal-2023-halueval} \\
          & \textbf{Dynamic:}            & \\
          & \quad Temporal               & \cite{chen-etal-2021b-timeqa,zhang-choi-2021-situatedqa,liska-etal-2022-streamingqa,kasai-etal-2022-realtimeqa,yu2024kola,liu-etal-2024-untangle} \\
          & \quad Data contamination     & \cite{livebench,tang2024evowiki,wu2024antileak} \\
          \cline{1-3}
        \noalign{\vskip 1ex}
        \multirow{8}{*}{\rotatebox{90}
        {\small\textbf{Reasoning}}}
          & \textbf{Math:}               & \\
          & \quad Primary school         & \cite{wang-etal-2017-deep,amini-etal-2019-mathqa,miao-etal-2020-diverse,cobbe2021training} \\
          & \quad College                & \cite{mathvista,arb,hendrycksmath2021} \\
          & \quad Olympiad               & \cite{omnimath,olympiadbench,frontiermath} \\
        \addlinespace[0.8ex]
          & \textbf{Coding:}             & \\
          & \quad Python                 & \cite{hendrycksapps2021,austin2021program,Lai2022DS1000,jimenez2024swebench,zhuo2024bigcodebench,effibench} \\
          & \quad Multi-lingual          & \cite{multiple,chen2021codex,codescope,mceval,livecodebench} \\
          & \quad Software Development   & \cite{DBLP:journals/corr/abs-2102-04664,codeeditorbench} \\
      \end{tabular}%
    }
    &
    {%
      \begin{tabular}[t]{@{}r@{\quad}l@{\quad}r@{}}
        \multirow{16}{*}{\rotatebox{90}{\small\textbf{Reasoning}}}
          & \textbf{Commonsense:}        & \\
          & \quad Resources              & \cite{cyc,opencyc,omcs,conceptnet} \\
          & \quad Social                 & \cite{naive,rocstory,sakaguchi2021winogrande,COPA,VCR,ecare} \\
          & \quad Temporal               & \cite{ZKNR19,udst,wang-etal-2022-maven} \\
          & \quad Physical               & \cite{PIQA,zellers2019hellaswag,acr,SWAG,talmor-etal-2019-commonsenseqa,Ordinal} \\
          & \quad Domain-Specific        & \cite{aser,mmekg,atomic,visualcomet} \\
        \addlinespace[0.8ex]
          & \textbf{Logic:}              & \\
          & \quad Deductive              & \cite{folio,ProofWriter,prontoqa,LogiQA,reclor} \\
          & \quad Inductive              & \cite{chollet2019measure,1darc,sygus,CLUTRR} \\
          & \quad Abductive              & \cite{acr,TrueDetective,AbductionRules} \\
        \addlinespace[0.8ex]
          & \textbf{Planning:}           & \\
          & \quad Task                   & \cite{gosc,wikihow,PlanBench,taskbench,NATURAL,workbench,ultratool} \\
          & \quad Agent                  & \cite{WebShop,TravelPlanner,smartplay,ontheplan,cantplan,saycan} \\
          & \quad Embodied               & \cite{textworld,Habitat,li2021igibson,alfred,BEHAVIOR-1K,alfworld} \\
        \addlinespace[0.8ex]
          & \textbf{Long context}        & \cite{an2023leval,bai2024longbench,zhang-etal-2024-bench,longbenchv2,li2024longcontextvsrag} \\\\
          & \textbf{Miscellaneous}       & \cite{wei2022chain,rahmanzadehgervi2024vision,spatial,calvin} \\
      \end{tabular}%
    }
    &
    {%
      \begin{tabular}[t]{@{}r@{\quad}l@{\quad}r@{}}
        \multirow{10}{*}{\rotatebox{90}{\small\textbf{Instruction Following}}}
          & \textbf{Task-based}         & \cite{radford2019language,mishra-etal-2022-cross,sanh2022multitask,chia-etal-2024-instructeval,weifinetuned,wang-etal-2022-super} \\\\
          & \textbf{Real-World Prompts} & \cite{eccleston2023sharegpt,conover2023free,openassistant,zheng2023judging} \\\\
          & \textbf{Automated}          & \cite{alpaca_eval,vicuna2023,li2024crowdsourced,zheng2023judging} \\\\
          & \textbf{Style-Control}      & \cite{li2024stylecontrol,chen-etal-2024-humans,dubois2024length,park-etal-2024-disentangling} \\\\
          & \textbf{Constraint-based}   & \cite{zhou2023instruction,jiang2023followbench,lin2024wildbench} \\\\
          \cline{1-3}
        \noalign{\vskip 1ex}
        \multirow{8}{*}{\rotatebox{90}{\small\textbf{Safety}}}
          & \textbf{Content Safety}       & \cite{hartvigsen2022toxigen,gehman-etal-2020-realtoxicityprompts,lin-etal-2023-toxicchat,ji2024beavertails,sun-etal-2022-safety,fleisig-etal-2023-fairprism} \\\\
          & \textbf{Multi-Dimension}      & \cite{ganguli2023challenges,DecodingTrust,aegis,SORRY-Bench,xsafety,S-Eval,safetybench,SG-Bench,SALAD-Bench,CHiSafetyBench,bai2022training,hendrycksaligning,wang2024not,koh2024can,pang2024self} \\\\
          & \textbf{Adversarial Robustness}& \cite{raheja2024recent,wichers2024gradient,hardy2024astprompter,yang2024sop,yu2023gptfuzzer,zhang2024wordgame,advbench,shen2024anything,aart,AdvPromptSet,kour-etal-2023-unveiling,cpad,tedeschi2024alert,AnthropicRedTeam,bai2022training,xu-etal-2021-bot,liu2024autodan,zou2023universal,liu2025autodan-turbo} \\\\
          & \textbf{Agentic Safety}      & \cite{luo2025agrail,xiang2024guardagent,xu2024advweb,liao2024eia,shi2024ehragent,zhang2024agent,zhang2025agent,yuan2024r} \\
      \end{tabular}%
    }
    \\
    \bottomrule
  \end{tabular}
\end{table*}

\subsubsection{Adversarial Robustness}
In previous sections, we primarily focused on detecting whether models generated unsafe content. Benchmark datasets usually collect potentially harmful prompts through pattern matching or rule-based filtering to measure the toxicity probability of model outputs.
As LLMs advanced, researchers recognized that static prompts were insufficient to comprehensively expose risks. This led to a shift toward adversarial testing or red-teaming methods, where humans or automated algorithms iteratively refine prompts to bypass safeguards.
Before benchmarks, we first briefly introduce several typical attack methods as one of the basic evaluation components.
There are two groups of methods: white-box and black-box attacks~\cite{raheja2024recent}.
For white-box attack methods, Gradient-Based Red Teaming (GBRT)~\cite{wichers2024gradient} uses model gradients to optimize prompts that trigger policy violations. 
In contrast, many black-box strategies treat the model as an API and use search or another LLM to craft exploits. Recent methods include reinforcement learning to generate realistic but harmful queries~\cite{hardy2024astprompter} and persona-driven attacks, such as SoP~\cite{yang2024sop}, which creates multi-character role-play scenarios to exploit a model’s social compliance.
To highlight the importance of robust input processing, fuzzing techniques have been proposed to capture subtle prompt variations. GPTFuzz~\cite{yu2023gptfuzzer} mutates seed prompts and reveal sensitivity to slight input perturbations, while WordGame~\cite{zhang2024wordgame} conceals harmful requests behind scrambled text to bypass content filters. 
These diverse red-teaming approaches exploit different model vulnerabilities (from over-confidence and ``distractibility'' to context manipulation and timing), greatly expanding the adversarial toolkit.

Alongside attack methodologies, researchers have built evaluation datasets to benchmark LLM robustness under attack. We can roughly classify them into three groups. The first type adopts single-turn attack, similar with those introduced above yet with intentional design to induce the model into ignoring its safety guardrails (e.g., a universal adversarial suffix). Example datasets include AdvBench~\cite{advbench}, ForbiddenQuestions~\cite{shen2024anything}, AART~\cite{aart}, AdvPromptSet~\cite{AdvPromptSet}, AttaQ~\cite{kour-etal-2023-unveiling}, CPAD~\cite{cpad}, and ALERT~\cite{tedeschi2024alert} that introduces a fine-grained risk taxonomy consisting of 6 macro and 32 micro categories.
Second, datasets like AnthropicRedTeam~\cite{AnthropicRedTeam}, AnthropicHarmlessBase~\cite{bai2022training}, and Bot-Adversarial Dialogue (BAD) dataset~\cite{xu-etal-2021-bot} leverage human or agent to curate adversarial dialogues with the goal of exposing model failures in multi-turn interactions.
The third group of datasets aim to proactively spotting vulnerabilities of models. Recent literature deploys evolutionary red teaming processes to optimize attacks. A representative work of this kind is AutoDan~\cite{liu2024autodan} which employs an hierarchical genetic algorithm to evolve prompts.
Unlike previous attacks~\cite{zou2023universal} that require gradient-based optimization, AutoDan efficiently operates mutations and crossovers of attack prompts as paraphrasing and linguistic exchange of paragraph content, easily strengthening any manually designed attacks without losing their semantic meaningfulness.
The more recent follow-up AutoDan-Turbo~\cite{liu2025autodan-turbo} further extends such a genetic process to evolve high-level attacking strategies, leading to a life-long learnable red teaming system that can be generally applicable to discover unforeseen threats to forthcoming LLMs.

\subsubsection{Agentic Safety}~\label{sec:data_agentsafe}
The newest frontier in LLM safety evaluation is agentic safety, which assesses LLMs that act as autonomous agents, operating tools or navigating environments on behalf of users. These agents must not only avoid producing harmful content, but also avoid harmful actions.
This introduces new safety challenges rooted from both users and environments, which involve handling multifaceted threats associated with user authorities, system mechanisms and runtime user-system interaction sessions~\cite{luo2025agrail}.
In terms of the environment, many studies focus on web agents like Mind2Web-SC~\cite{xiang2024guardagent}, AdvWeb~\cite{xu2024advweb}, EIA~\cite{liao2024eia}.
To explore more domains, EICU-AC~\cite{shi2024ehragent} targets the medical domain to evaluate access control of LLM agents based on user authorization when processing electronic health records.
Safe-OS~\cite{luo2025agrail} evaluates the robustness of OS agents, meanwhile, investigates a broad range of attacks including prompt injection, system sabbotage attacks, and environment attacks.
Agent-SafetyBench~\cite{zhang2024agent} encompasses 349 interactive environments (simulated scenarios) with 2,000 total test cases, covering 8 categories of safety risks and 10 common failure modes for agent behavior.
ASB~\cite{zhang2025agent} includes 10 scenarios and benchmark various attack tools, e.g., prompt injection, memory poisoning, and backdoor.
Instead of building costly environments, R-judge~\cite{yuan2024r} consists of 569 logs of multi-turn agent interactions (drawn from various simulated applications) with annotated risk events covering 27 scenario types and 10 distinct risk categories. The task is for an LLM to read the log and correctly flag any unsafe decisions or outcomes.

Clearly, agentic safety evaluation is inherently more challenging than static LLM evaluation, because it requires simulating an interactive environment. Besides, the benchmark has to define the risk taxonomy, evaluator for open-ended responses or actions, attack tools for robustness assessment, etc.
Therefore, in the future more environments are expected to cover various domains.
In these simulations, as agents are intended to handle long-horizon tasks, another evaluation focus is long-term robustness under distribution shift. An agent might start aligned, but after many steps or after successively encountering adversarial inputs, it could deviate from policy.
Finally, a critical aspect of agentic safety is balancing utility with safety. If an agent is overly constrained, it may refuse to use its tools at all or become useless.

\subsection{Integrated Capabilities: General-purpose Evaluation}

Early benchmarks for LLMs often targeted isolated capabilities, e.g., logical reasoning or instruction following in separate tests. However, real-world tasks rarely exercise these skills in isolation.
Recent evaluation efforts therefore emphasize integrated assessment from GLUE to MMLU, and to Big-Bench, measuring how well an LLM can combine knowledge, reasoning, instruction following, multi-modal understanding, and safety together. 
The goal is to move beyond concrete ability tests toward holistic evaluation for general-purpose performance or artificial general intelligence,  mirroring the integrated demands of real-world use.
After the discussion on each ability, we summarize their taxonomy in Figure~\ref{fig:data_tax}. Now, we will discuss the overlaps of integrated or comprehensive benchmarks.

\subsubsection{Interplay Among Evaluation Capabilities}
In practice, while we have detailed the evaluation datasets and potential challenges for each individual capability in preceding sections, these capabilities are inherently intertwined. In this section, rather than trivially enumerating every possible combination, we will instead examine selected examples of capability interplay and then discuss generalized and holistic evaluation.

\textbf{Knowledge \& reasoning.}
Knowledge and reasoning are inherently intertwined. Effective reasoning fundamentally depends on a model’s underlying knowledge base. As illustrated in earlier sections on mathe reasoning, it assumes mastery of basic mathematical theorems; coding reasoning requires knowledge of programming languages; commonsense reasoning often evaluates familiarity with commonsense knowledge; planning necessitates procedural knowledge; even purely logical reasoning rarely operates in complete isolation from knowledge --- even using arbitrary symbols or ``neutral'' facts to eliminate the influence of memorized world knowledge, it may risk oversimplifying language understanding and introduce artificial patterns that models exploit without generalizing to real-world reasoning~\cite{liu2025logical}.
Conversely, knowledge-intensive tasks frequently demand reasoning capabilities. For example, open-domain QA datasets like HotpotQA~\cite{yang2018hotpotqa} and StrategyQA~\cite{geva2021did} usually require models to retrieve and interconnect multiple facts before reaching conclusions, i.e., multi-hop reasoning.

\textbf{Instruction following \& Knowledge \& Reasoning.}
Broadly speaking, instructions, as user inputs to models, can encompass any task description. This implies that knowledge and reasoning capabilities can also be considered sub-skills of instruction following.
For example, in practice, users often ask LLMs to perform multi-step tasks via natural language instructions, ``Explain how to solve this math problem step by step.'' or ``Analyze the argument in the following paragraph.''
Benchmarks like MT-Bench~\cite{zheng2023judging} combine instruction following with knowledge, requiring LLMs to generate responses across diverse topics while adhering to user directives. Similarly, VicunaEval~\cite{vicuna2023} integrates inherent knowledge with precise instruction execution.
Reasoning tasks can also be framed as instructions. They usually define a set of rules, where LLMs must follow logical steps to derive final states from initial conditions~\cite{wei2022chain}. Relevant benchmarks also include CodeXGlue~\cite{DBLP:journals/corr/abs-2102-04664} and BigCodeBench~\cite{zhuo2024bigcodebench}. These require LLMs to follow detailed programming instructions for tasks like code completion, unit testing, and documentation generation.

\textbf{Interaction with multi-modal understanding.}
Multi-modal understanding is inherently orthogonal to other capabilities. All previously discussed evaluation benchmarks can be extended to additional modalities.
A clear example is VQA benchmarks, where models are given an image and a related question. To succeed, the model must interpret visual content (detect objects, scenes, text in the image) and often use world knowledge or reasoning to answer the question.
Benchmarks like OK-VQA~\cite{marino2019ok} specifically target this intersection, requiring models to integrate external knowledge with visual comprehension.
For multi-modal reasoning, specialized benchmarks emerge.
Math-Vista~\cite{mathvista} and MathVision~\cite{mathvision} targets the evaluation of multi-modal math reasoning by combining diagrammatic representations with textual problem statements.
MMMU~\cite{yue2023mmmu} and MMMU-Pro~\cite{yue2024mmmupro} presents college-level questions that interleave text with heterogeneous visual inputs, demanding both domain-specific knowledge and advanced reasoning skills. This dataset pushes models to draw upon a broad base of subject knowledge while performing deliberate, expert-level reasoning across multiple disciplines.
OlympiadBench~\cite{olympiadbench} further pushes difficulty to Olympiad-level in math and physics context.
Similarly, visual instruction tuning~\cite{liu2023visual,xu2024vision} bridges visual understanding with instruction following, while Huang et al.\cite{huang2023visual} provide a comprehensive survey.
In summary, multi-modal evaluation expands the scope of integrated assessment. It ensures LLMs’ general capabilities extends beyond text to interpret and reason about visual (or auditory, etc.) worlds in conjunction with language.

\textbf{Interaction with safety.}
The safety capabilities of LLMs are also orthogonal to other competencies yet critically important for real-world deployment. Increasingly, benchmarks incorporate safety evaluations alongside knowledge and reasoning tasks. For instance, TruthfulQA~\cite{lin2021truthfulqa} systematically tests models with 818 challenging questions spanning 38 domains (e.g., health, law, finance) to distinguish between truthful responses and fluent but factually incorrect answers that mimic human plausibility. This paradigm evaluates not just factual knowledge and linguistic proficiency, but crucially measures truthfulness alignment, prioritizing correct, honest responses over eloquently stated misconceptions and thereby integrating factual reasoning with safety metrics.
For intersection with instruction-following capabilities, models that unconditionally obey user requests risk generating harmful outputs. Effective safety alignment necessitates the ability to override instructions when appropriate. Contemporary evaluations address this by incorporating refusal-worthy prompts (Section~\ref{sec:con_safe}), where properly aligned models must demonstrate refusal competence or safe response redirection.
Notably, SafeBench~\cite{ying2024safebench} provides a systematic framework for evaluating multi-modal LLM safety, extending these principles to complex, real-world deployment scenarios. This comprehensive approach ensures that safety mechanisms remain robust when models operate at the intersection of knowledge retrieval, reasoning, and instruction execution, a critical requirement for trustworthy AI systems.

\subsubsection{Comprehensive Evaluation}
Based on the above analysis, the field is progressively integrating knowledge, reasoning, instruction following, multimodal understanding, and safety into integrated benchmark suites, moving beyond isolated skill testing, for a comprehensive measure of a model’s general-purpose capabilities. This is not only the abilities are inherently intertwined, but also real-world deployment requires the simultaneous application of these capabilities. 
Early initiatives like GLUE~\cite{wang-etal-2018-glue} and SuperGLUE~\cite{wang2019superglue} pioneered this approach by aggregating several common NLP tasks, enabling multifaceted evaluation of pre-trained language models like BERT during fine-tuning. 
Subsequent comprehensive benchmarks expanded the scope. 
MMLU~\cite{hendrycks2020measuring} includes 57 subjects, including elementary mathematics, US history, computer science, and law. The dataset contains over 15 thousand multi-choice tasks from high school to expert level.
MMLU-Pro~\cite{wang2024mmlu} updates the MMLU framework with more challenging reasoning tasks, enhanced robustness, and reduced dataset noise.
As the transition from task-oriented to capability-centric evaluation occurs, BIG-bench~\cite{bigbench} curates over 200 diverse tasks covering mathematics, linguistics, commonsense reasoning, and social bias analysis among others. To address computational constraints, BIG-bench Lite provides a distilled 24-task subset for efficient performance measurement.
Building on that idea, recent benchmark collections like HELM~\cite{liangholistic} and VHELM~\cite{lee2024vhelm} explicitly report a profile of each model across many aspects, from accuracy on academic questions to robustness under input perturbations and fairness in responses. The aim is to identify not just ``which model is best'' but in what ways a model is strong or weak, and how well it balances the competing demands of capability and alignment. 

Another holistic evaluation frameworks put LLMs into agent roles, asking them to operate in interactive environments or multi-step decision problems, which tests a convergence of capabilities in scenarios closer to real-world deployment.
Except for prior agent planning (Section~\ref{sec:data_plan}) and agentic safety(Section~\ref{sec:data_agentsafe}), there are also some general evaluation involving one or multiple LLM agents to collaborate or compete.
LLMs-as-an-Examiner~\cite{bai2023benchmarking} simulates peer review assessments. Each LLM operates like an examiner to generate queries and also judge the responses from other LLMs. By collaboratively determining the evaluation results, this process reduces biases and enhances fairness in evaluations.
Auto-Arena~\cite{zhao2024autoarenaautomatingllmevaluations} introduces discussion among LLM agents to automate this process.
Inspired by educational assessment processes, AutoDetect~\cite{autodetect} employs three LLM-powered agents --- Examiner, Questioner, and Assessor --- that collaborate to generate test scenarios and analyze model responses. 
Agent-CQ~\cite{Agent-CQ} leverages LLMs to automate the creation and assessment of clarifying questions in conversational search systems,
while \textsc{LegalAgent}~\cite{li2024legalagentbench} pushes the evaluation further to agent-based legal reasoning.
There are also visual agent evaluation framework to evaluate the ability of MLLMs to perform complex real-world tasks like UI operation on mobile devices~\cite{zhang2024android}, robotic control in household tasks~\cite{shridhar2020alfred}, card-based games~\cite{zhai2024fine}, and navigation~\cite{wang2024navigating,anderson2018vision,qi2020reverie}.
Other similar frameworks include IQA-Eval~\cite{iqaeval}, ALI-Agent~\cite{ali_agent}, ChatEVal~\cite{chateval}, MATEval~\cite{li2024mateval}, and AgentSims~\cite{agentsims}.
Unlike static datasets, this type of interactive benchmarks also test adaptability and decision-making. An agent can observe new information and must decide its next action. More practically, a model might initially answer a question incorrectly, but in an interactive setting it could be given feedback or detect the error and correct itself. Metrics for such evaluations can include success rate, efficiency (steps taken), and qualitative ratings of the agent’s behavior.

\section{Auto-evaluation}
\label{sec:autoeval}

In Section~\ref{sec:bench}, we introduced commonly used datasets categorized by five core capabilities and discussed their interplay. However, these static datasets lead to delayed updates of test sets, hindering their alignment with model progress. Additionally, they are still susceptible to performance overestimation due to data contamination. In this section, we first introduce several dynamic benchmarks and live leaderboards, followed by methods for automated dataset curation and evaluation.
The Github page we will maintain and welcome any collaborators is \url{https://github.com/ALEX-nlp/Chapter3_Awesome_Paper_List}.

\subsection{Dynamic benchmarks}
\label{sec:dybench}
Dynamic benchmarks aim at continuously updating the testing data to offer a fairer assessment. There are mainly two types of advantages.
First, it considers the dynamic nature of world knowledge, thus preventing false negatives caused by outdated information and accurately assessing model latest advancements. 
Early works highlight the timeliness of knowledge by introducing timestamp, where a piece of knowledge holds true only within its own timestamps. To obtain accurate timestamp, common sources for dataset curation include WIKIDATA~\cite{chen-etal-2021b-timeqa}, news articles~\cite{liska-etal-2022-streamingqa}, or existing datasets annotated by crowd-sourcing workers~\cite{zhang-choi-2021-situatedqa}. The target of such evaluation is similar with temporal commonsense reasoning in Section~\ref{sec:common_reason}. 
Building on top of them, recent benchmarks target real-time evaluations. 
\textsc{REALTIMEQA}~\cite{kasai-etal-2022-realtimeqa} evaluates models weekly on approximately 30 multiple-choice questions derived from recent news events.
This benchmark highlights the importance of continual learning and real-time knowledge integration for accurate and timely responses.
\textsc{Kola}~\cite{yu2024kola} and \textsc{Knot}~\cite{liu-etal-2024-untangle} take a step further by not only evaluating the coverage of the rapidly changing world knowledge but also the ability of models to integrate the new knowledge with the existing knowledge.

The second advantage of dynamic benchmarks is that consistently updating data can mitigate the data contamination issue. There are two type of methods. The first group still leverages the timeliness of knowledge. Since as long as the evaluation data pertains to the latest knowledge, the model would not have been exposed to it, thereby no testing data shall be seen during training. However, other researchers also point out that the risk lies in the difficulty of completely distinguishing new from old knowledge based on a specific cutoff date (such as the model's release date). For instance, even if a movie is released after this date and the model should not have seen it, necessary information might have already been exposed to the model through early promotions and related activities~\cite{tang2024evowiki}.
Representative works include EvoWiki~\cite{tang2024evowiki} and AntiLeak-Bench~\cite{wu2024antileak}. For example, EvoWiki is an evolving dataset that categorizes information into stable, evolved, and uncharted states. By comparing the information before and after the LLM's cut-off date, stable data do not change and evolved data indicate an update. While, uncharted data only involve those events happened thereafter.

Instead of the timeliness of knowledge, another line of research highlights the dynamics of data --- as long as the evaluation data keeps changing, they are hardly exposed to the model during evaluation.
Dynabench~\cite{dynabench} is an open-source platform that incorporates human-and-model-in-the-loop dataset creation. Unlike traditional static benchmarks, Dynabench enables annotators to craft examples that challenge current models, revealing their weaknesses and promoting the development of more robust systems. This dynamic approach directly integrates human feedback into the evaluation process. 
Livebench~\cite{livebench} aims at releasing new questions monthly, sourced from recent information such as math competitions, arXiv papers, and news articles, thereby ensuring that models are assessed on fresh, unseen data. It encompasses six categories: math, coding, reasoning, language comprehension, instruction following, and data analysis, each with tasks that have verifiable ground-truth answers.
Focusing on specific coding task, Livecodebench~\cite{livecodebench} instead continuously collects new problems from coding competitions on platforms like LeetCode, AtCoder, and CodeForces. Beyond code generation, LiveCodeBench also assesses capabilities in self-repair, code execution, and test output prediction, providing a holistic view of an LLM’s coding proficiency.

\textit{Live Leaderboard}
To facilitate a convenient evaluation, some researchers develop and maintain real-time evaluation platforms that are updated either manually or automatically.
Chatbot Arena~\cite{chatbot_arena} is an open evaluation platform based on human preferences. Users can engage in side-by-side conversations with anonymous AI models and vote for their preferred responses, facilitating direct comparisons of AI capabilities in real-world scenarios. The platform employs the Elo rating system to rank models based on user votes. Since its launch in May 2023, Chatbot Arena has attracted millions of participants and collected over 800,000 votes, becoming a critical resource for live, community-driven LLM evaluation.

EvalPlus Leaderboard~\cite{evalplus} is a platform for code generation. It utilizes enhanced benchmarks, such as HumanEval+ and MBPP+, which offer significantly more test cases than their original versions, to assess models’ code correctness and efficiency. By ranking models based on metrics like pass@1 using greedy decoding, the leaderboard provides insights into each model’s coding proficiency and robustness.

Open LLM Leaderboard~\cite{openllm} tracks, ranks, and evaluates open-source LLMs and chatbots. It provides a centralized resource for comparing the performance of various models across multiple benchmarks, facilitating informed decisions for researchers and developers in the AI community. Users can submit their models for automated evaluation on Hugging Face’s GPU cluster, ensuring standardized assessments. The leaderboard is continuously updated, reflecting the latest advancements.

C-Eval Leaderboard~\cite{C-eval} is a comprehensive Chinese evaluation suite, consisting of 13,948 multiple-choice questions spanning 52 diverse disciplines, including humanities, science, and engineering, and is structured across four difficulty levels: middle school, high school, college, and professional. Notably, C-Eval includes a challenging subset known as C-Eval Hard, which focuses on subjects requiring advanced reasoning skills, such as advanced mathematics and college physics.

Clearly, to ensure the high quality of dynamic benchmarks, the key is how to automate dataset curation and evaluation, which will be detailed in the next section.

\subsection{Automated Dataset Curation}
\label{sec:autodata}

Qualified human-annotated data requires substantial budgets and time cost, thus being particularly vulnerable to the rapid outdatedness and potential information leakage. Accordingly, more and more evaluation datasets are constructed in auto-synthesized manners. In this section, we summarize the common auto-synthesis strategies into three main branches: compilation, derivation and generation. (1) \textit{Compilation} involves combining or selecting existing annotations to align with the intended use of the dataset. (2) \textit{Derivation} utilizes existing datasets but modifies annotations or adds new components to serve specific purposes. (3) \textit{Generation} involves partially or completely constructing datasets by automatically generating new contexts or annotations, often with the assistance of LLMs. We detail these three strategies as below. Note that the strategies summarized here are not exclusive: the construction of one evaluated dataset can leverage multiple strategies.

\subsubsection{Compilation} Compilation is the most simple and widely-used approach for building a new evaluation dataset. It can be further divided into Combination and Selection.

\noindent \textbf{Combination} integrates existing annotations into a new, single benchmark. These benchmarks are designed to evaluate the general capabilities of LLMs~\cite{wang-etal-2018-glue, wang2019superglue, bigbench, livebench, ying2024mmt}, or assess specific abilities in a more comprehensive and robust manner like math/STEM~\cite{hendrycksmath2021, chen-etal-2023-theoremqa, wang2024scibench, olympiadbench}, coding~\cite{jimenez2024swebench, livecodebench}, long-context understanding~\cite{bai2024longbench, an2023leval, song2024milebench}, information retrieval~\cite{thakur2021beir, muennighoff-etal-2023-mteb, faysse2024colpaliefficientdocumentretrieval, ying2024intuitive}, etc.
The biggest challenge for combination is how to construct the benchmark a hierarchical and reasonable taxonomy. The taxonomy in existing benchmarks are usually designed in the following dimensions: 

\begin{itemize}
    \item Target abilities. For example, MathVista~\cite{mathvista} summarizes seven mathematical reasoning capabilities. MMBench~\cite{MMBench} designs a three-level, twenty-subclass taxonomy tree to evaluate the perception and reasoning abilities of LVLMs. Recently, MEGA-Bench~\cite{chen2024megabenchscalingmultimodalevaluation} includes over 500 real-world tasks within the hierarchical taxonomy.
    \item Discipline and/or difficulty. Most benchmarks sourced from examinations or exercises~\cite{hendrycks2020measuring, C-eval, zhang2023m3exam, yue2023mmmu, chen-etal-2023-theoremqa} are usually categorized by disciplines and/or difficulties. For instance, MMLU~\cite{hendrycks2020measuring} and MMMU~\cite{yue2023mmmu} incorporate questions from 57 and 30 subjects, respectively. MATH~\cite{hendrycksmath2021} and M3Exam~\cite{zhang2023m3exam} divide their questions into 5 and 3 difficulty levels. 
\end{itemize}

Given the critical importance and relatively limited scale of the taxonomy, it still heavily relies on human design. Traditionally, the taxonomy is entirely designed by the main contributors of the benchmarks. However, as the taxonomy scale expands, the responsibility for its expansion is distributed. For example, BIG-Bench~\cite{bigbench} encourages the entire community to submit pull requests for new tasks. MEGA-Bench~\cite{chen2024megabenchscalingmultimodalevaluation} initially provides a draft two-level taxonomy and invites all project members to contribute to its growth.
When the taxonomy is completed, they either collect previous datasets (such as GLUE~\cite{wang-etal-2018-glue}, SuperGLUE~\cite{wang2019superglue}, BIG-bench~\cite{bigbench}, and MathVista~\cite{mathvista}) or gather annotations from multiple sources like websites, textbooks, or real-world data (such as MMLU~\cite{hendrycks2020measuring}, MATH~\cite{hendrycksmath2021}, OlympiadBench~\cite{olympiadbench}, C-Eval~\cite{C-eval}, M3Exam~\cite{zhang2023m3exam} and MMMU~\cite{yue2023mmmu}).

\noindent \textbf{Selection} involves filtering annotations to create new benchmarks. Such process usually serves for three purposes:

\begin{itemize}
    \item Scale control. Some benchmarks, especially those constructed in combination way, sample annotations randomly from previous datasets to control the scale of the curated datasets~\cite{wang2024mintevaluatingllmsmultiturn, bai2024longbench}.
    \item Preliminary filtering. The raw data crawled from real-world are sometimes noisy. It requires a preliminary selection to improve the recall of qualified data with the minimum time and budget cost. Simple but effective heuristic rules are usually adopted. For example, ELI5~\cite{fan-etal-2019-eli5} and BRIGHT~\cite{su2024brightrealisticchallengingbenchmark} select high-quality posts and/or answers (measured by views, votes, URL numbers) in the Reddit or StackExchange. Benchmarks for tabular task like HybridQA~\cite{chen-etal-2020-hybridqa} and FinQA~\cite{chen-etal-2021-finqa} retain moderate-size tables (measured by row and column numbers) to balance the information amount and task difficulty. SciToolBench~\cite{ma-etal-2024-sciagent} picks out tools (Python functions) passing the unit tests to ensure the correctness of these tools.
    \item Post-refinement. When the annotations have been made from scratch or collected from previous datasets, an additional selection (after the main construction process) further benefit the dataset from various aspects like quality, diversity, difficulty, etc. Compared to pre-filtering, it requires more customized assessment and often involves LLMs/LVLMs as judges. For instance, MMLongBench-Doc~\cite{ma2024mmlongbenchdocbenchmarkinglongcontextdocument} evaluates the document understanding abilities of LVLMs (instead of their intrinsic knowledge) and thus employs GPT-4o to remove sample candidates which can be directly answered without the access to documents. VisRAG-Bench~\cite{yu2024visragvisionbasedretrievalaugmentedgeneration} introduces Llama-3 to filter out context-dependent queries which are not appropriate for open-domain retrieval task. MMStar~\cite{chen2024we} evaluates the visual mathematical reasoning abilities and thus employs GPT-4 to remove sample candidates which can be answered by text-only information. HaluEval~\cite{li-etal-2023-halueval} and MMHal-Bench~\cite{sun2023aligninglargemultimodalmodels} introduce ChatGPT and LLaVA to rank and select high-quality hallucinated answers (\ie more plausible and close to the correct answers) from previous generations. For a simple approach to achieve strong reasoning during test-time inference, S1~\cite{muennighoff2025s1} has conducted post-refinements regarding quality, difficulty, and diversity. For quality, they filter out low-quality examples by checking if they contain any string patterns with formatting issues, such as ASCII art diagrams, non-existent image references, or inconsistent question numbering. For difficulty, they first use Claude3.5 to select the correct responses, and then measure the problem difficulty through the token length of each response including the reasoning process. For diversity, they introduce the Mathematics Subject Classification (MSC) system (e.g., geometry, dynamic systems, real analysis, etc.) and classify each question into these specific domains using Claude 3.5 Sonnet, keeping balanced distribution across different domains.
\end{itemize}

\subsubsection{Derivation} Derivation is an automatic construction strategy somewhat between compilation and generation. It still heavily relies on existing annotations, but introduces significant modifications. We further categorize derivation into two subtypes, Transfer and Supplementary, as detailed below.

\noindent \textbf{Transfer} usually occurs when we evaluate the identical or highly-similar tasks/abilities under different settings. In such cases, creating new benchmarks from scratch is neither necessary nor economical. The common choice is to make new benchmarks by transferring from an existing, well-developed ones. 
For similar tasks, BEIR~\cite{thakur2021beir} and ViDoRE~\cite{faysse2024colpaliefficientdocumentretrieval} are information retrieval benchmarks collected from multiple QA datasets by simple conversion: (i) from question to query. (ii) merge passages as retrieval corpus. Being an open-domain QA dataset, OTT-QA~\cite{chen2021ottqa} rewrites the queries in HybridQA~\cite{chen-etal-2020-hybridqa} for decontextulization. To explore the long-context/multi-page settings, MP-DocVQA~\cite{tito2023hierarchicalmultimodaltransformersmultipage} and LongBench~\cite{bai2024longbench} increase the document lengths in previous datasets like DocVQA~\cite{docvqa} by incorporating additional context pages or similar paragraphs. 
For the same tasks at different modalities, with the development of LVLMs, benchmarks for many critical capabilities and practical task in text domains are converted to visual domains and used to evaluate LVLMs. For example, Multimodal-Mind2Web~\cite{zheng2024seeact} and SWE-bench Multimodal~\cite{yang2024swebenchmultimodalaisystems} trace back the webpage screenshots which are used as textual format in Mind2Web~\cite{deng2023mind2web} and SWE-bench~\cite{jimenez2024swebench} to evaluate the agent and coding abilities of LVLMs. Similarly, Wiki-VISA~\cite{ma2024visaretrievalaugmentedgeneration} and M3DocVQA~\cite{cho2024m3docragmultimodalretrievalneed} render the Wikipedia URLs in NQ~\cite{kwiatkowski-etal-2019-natural} and MultimodalQA~\cite{talmor2021multimodalqa} datasets towards the evaluation of visualized document understanding and grounding.

\noindent \textbf{Supplementary} usually occurs when we have the benchmark about some tasks and aim to evaluate their further/successive tasks. In such cases, it is a natural choice to build corresponding benchmarks in supplementary approach, \ie adding new annotations based on the original ones. To explore whether retrieval benefits code generation
models, CodeRAG-Bench~\cite{wang2024coderagbenchretrievalaugmentcode} is derived from coding benchmarks~\cite{chen2021codex, livecodebench} by augmenting collected documents as retrieval corpus. HellaSwag~\cite{zellers2019hellaswag} is synthesized from ActivityNet~\cite{Heilbron2015ActivityNetAL} as a QA dataset in which the negative choices are adversarially synthesized. RuleBench~\cite{sun2024instructionfollowingevaluatinginferential} induces rules from multiple logic-related datasets and add these rules to form a new benchmark for inferential rule-following evaluation. MMLongBench-Doc~\cite{ma2024mmlongbenchdocbenchmarkinglongcontextdocument} and MuirBench~\cite{wang2024muirbenchcomprehensivebenchmarkrobust} add unanswerable ones by replacing keywords in original questions, thereby detecting potential hallucinations. Upon annotated events, MAVEN-FACT~\cite{li-etal-2024-maven} automatically generate their factualities for Event Factuality Detection (EFD) task.

\subsubsection{Generation}~\label{sec:autodata_generation}
When there are no corresponding qualified annotations for reuse and/or edition, generation becomes the indispensable choice for automatic data construction.

\textbf{Rule-based Generation} is still widely-used in the era of LLMs due to its efficiency and deterministic, especially under the following scenarios. (1) \textit{Mine real-world data}. The data for some tasks like math, coding and knowledge base are entailed in related informative platforms and knowledge-rich sources. It is natural to design automatic pipelines and extract these real-world high-quality data as evaluation benchmark. For example, LeanDojo~\cite{yang2023leandojo} and SWE-Bench~\cite{jimenez2024swebench}  extract proofs from Lean and pull requests from GitHub repositories, respectively. RealTimeQA~\cite{kasai2023realtime} extracts questions from news websites which requires latest knowledge. AntiLeak-Bench~\cite{wu2024antileak} leverages the knowledge among entities from Wikidata and synthesizes QA pairs for contamination-free evaluation. Recently, CODEELO~\cite{quan2025codeelobenchmarkingcompetitionlevelcode} extracts updated coding problems from CodeForces. (2) \textit{Synthesize for certain capability evaluation.} To evaluate some specialized capabilities of LLMs/LVLMs, it is also beneficial to create somewhat artificial but targeted datasets. To assess the comprehensive reasoning abilities across different modalities, MultimodalQA~\cite{talmor2021multimodalqa} creates cross-modal questions from single-modal questions by pre-defined, compositional templates. Similarly, RuleTaker~\cite{clark2020transformerssoftreasonerslanguage} generates facts and rules in logic, performs forward inference to derive all its implications, and obtains questions expressed in (synthetic) English using simple natural language templates. MM-NIAH~\cite{wang2024needle} concatenates interleaved image-text sequences from the OBELICS~\cite{laurencon2023obelics} dataset to create long-context documents, referred to as multimodal haystacks. POPE~\cite{li-etal-2023-evaluating} employs templates which convert image instances with object detection annotations to QA pairs for object hallucination evaluation.

\textbf{LLM-based Generation} has been an important approach and research topic for automated dataset construction. The motivation of LLM-based generation can be categorized into four aspects. 
(1) Label generation. Here the raw corpus already exists. An LLM replaces human annotators by producing labels, rationales, or exemplar responses.
Example methods cover multiple scenarios like role-playing~\cite{wang-etal-2024-rolellm}, multi-agent communication~\cite{CAMEL}, multi-turn interaction~\cite{wang2024mintevaluatingllmsmultiturn, bai-etal-2024-mt}, code generation~\cite{zhuo2024bigcodebench, gu2024cruxevalbenchmarkcodereasoning}, tool-use~\cite{qin2023toolllm}, \etc.
(2) Context generation. This aims to complete the missing parts of existing benchmarks, e.g., generating responses or options in multi-choice questions. In such cases, the generations are exactly the evaluation targets of the benchmarks. For example, HaluEval~\cite{li-etal-2023-halueval} and FavaBench~\cite{mishra2024finegrained} focus on hallucination evaluation and thus leverages ChatGPT to generate contexts with potential errors. A series of benchmarks~\cite{liu2024rmbenchbenchmarkingrewardmodels, li2024vlrewardbenchchallengingbenchmarkvisionlanguage, jin2024ragrewardbenchbenchmarkingrewardmodels,zhu2024unraveling} focus on reward model evaluation and therefore generate a pair/group of LLM-generated responses which are fed to reward model.
(3) Reference-based revision. This type of methods aim at generating new data based on provided reference. Representative works include WizardLM~\cite{xu2024wizardlm} and follow‑up works. They treat an existing dataset as scaffolding and issue editing instructions: add constraints, deepen reasoning, inject noise, or rephrase, to an LLM, which rewrites each item into harder variants. Ying et al.~\cite{ying2024automatingdatasetupdatesreliable} proposed two types of methods, mimicking and extending, to systematically update test sets to mitigate possible data contamination issue. This approach preserves topical relevance yet upgrades difficulty and coverage, functioning as controllable ``data augmentation for evaluation''.
(4) From-scratch generation. When no suitable seed corpus exists, carefully designed prompts elicit an LLM to invent both tasks and solutions. Self‑Instruct~\cite{wang2022self} pioneers this method by supplying a handful of seed exemplars and letting the model extrapolate thousands of similar instruction–response pairs.
LLM‑as‑Examiner~\cite{bai2024benchmarking} considers both evaluation breadth and depth to generate diverse evaluation data. For breadth of knowledge, they collects thousands of domain descriptions as instructions. For depth, they prompt LLMs to generate follow-up questions as well as the responses.
Note that the first three generation approaches mentioned above can also be viewed as data synthesis methods --- they reprocess existing data to maintain quality while ensuring flexibility.

\subsection{Pipeline of Automated Dataset Curation}
To conclude, towards a qualified benchmark, current works carefully consider and design the following steps: (1) Well-defined taxonomy. Under each topic and/or task type, personalized generation strategies or instructions are adopted by LLMs and significantly improves the coverage and quality of LLM generations (See more details about the taxonomy construction in previous discussion). For example, \cite{agentbench, li2023seedbenchbenchmarkingmultimodalllms, ying2024mmt, chen2024quantifying} categorize multiple tasks and feed task-specific instructions to LLMs for generate more high-quality QA pairs. (2) Step decomposition. The auto-annotation of benchmark is usually decomposed into several subsequent steps. For instance, many QA-formatted benchmarks~\cite{CAMEL, jiao-etal-2023-instruct, qin2023toolllm, jin2024ragrewardbenchbenchmarkingrewardmodels} separately generate the questions/instructions, answers/responses. Regarding more complicated task, \cite{laban-etal-2024-summary} synthesizes instances for \textit{Summary of Haystack} task by four sub-steps: insight generation, document generation, query generation and summary generation.  (3) Prompt strategy. Most earlier benchmarks~\cite{qin2023toolllm, ma-etal-2024-sciagent} draw inspirations from In-context Learning (ICL; \cite{dong2024surveyincontextlearning}) and provide seed examples in prompts. These examples explicitly instruct LLMs to generate annotations with desired contents and formats. Moreover, more detailed prompts, higher quality the generation content, making prompts written in detailed instructions become more and more popular~\cite{li2023seedbenchbenchmarkingmultimodalllms, ma2024visaretrievalaugmentedgeneration}. (4) Verification. The preliminary generation from LLMs shall undergo verification procedure before use. For math and coding tasks, the easiest verification occurs when ground-truth answers are known and the generation result are deterministic or executable. In such cases, rule-based parsers or programming executor~\cite{ma-etal-2024-sciagent, yue2023mammothbuildingmathgeneralist, zhang2024sciinstructselfreflectiveinstructionannotated} are adopted for verification. Also, LLM-as-a-judge evaluators can efficiently assess the generation quality (More details in the coming section).

\subsection{Evaluator}

As LLMs have unified a variety of natural language processing tasks through natural language generation, a significant shift has occurred in how open-ended responses are evaluated. Traditionally, evaluation relied on task-specific metrics, calculated by directly comparing model outputs to reference texts. For example, in classification tasks~\cite{sokolova2009systematic}, metrics such as accuracy, precision, recall, and F1 score are commonly employed; in ranking tasks~\cite{wang2013theoretical}, metrics like NDCG are typically used. Similarly, a similar approach was adopted for natural language generation tasks. BLEU~\cite{papineni2002bleu}, for instance, is an automatic metric that measures the quality of machine-generated translations by calculating the overlap of n-grams between the output and reference texts. A higher overlap indicates better alignment with the reference, suggesting higher translation quality. Likewise, ROUGE~\cite{lin2004rouge} evaluates the quality of summaries in the same way by calculating word overlap between the evaluated summary and human-generated ideal summaries. However, BLEU and ROUGE primarily rely on lexical matching, often overlooking lexical order and meaning. METEOR~\cite{banerjee2005meteor} improves upon BLEU by not only considering unigram overlap but also incorporating lexical stem and semantic matching, which better captures linguistic diversity in translations. Additionally, METEOR accounts for recall and the ordering of lexical matches, allowing for a more accurate assessment of machine translation quality. Despite these improvements, such metrics still heavily rely on surface-level lexical overlap, which often fails to capture deeper semantic nuances, coherence, or logical consistency. 

In response, many embedding-based evaluation methods have emerged to assess model-generated responses at the semantic level. BERTScore~\cite{zhang2019bertscore}, for example, evaluates the quality of machine-generated text by comparing the semantic similarity between the generated text and reference using contextual embeddings from BERT. Unlike traditional metrics that focus solely on lexical overlap, BERTScore captures deeper meaning by comparing the cosine similarity of token embeddings, which makes it more effective in handling synonyms, paraphrasing, and variations in sentence structure. However, BERTScore still heavily relies on the availability of reference answers. In evaluation tasks where references are scarce or difficult to obtain, this dependence significantly limits its applicability. Furthermore, while BERTScore shifts from word-level to semantic-level evaluation, it struggles to capture aspects beyond semantics such as helpfulness and harmlessness. These limitations have highlighted the need for the development of reference-free and multi-aspect evaluation methods.

As the performance of LLM has progressively advanced, GPT and other models are introduced to replace BERT as evaluators with or without reference, a.k.a., the proposal of the LLM-as-a-judge concept~\cite{zheng2023judging}. Moreover, thanks to the strong knowledge memorization and instruction following capabilities, LLM-as-a-judge can even evaluate the responses from multiple aspects, like informative, engaging, etc.
A typical example is GPTScore~\cite{fu2023gptscore}, which utilizes LLMs (including GPT-3~\cite{brown2020language}, OPT~\cite{zhang2022opt}, and FLAN~\cite{chung2024scaling}) to evaluate text quality across multiple aspects without relying on reference responses. 

Currently, the development of LLM-as-a-judge is continually progressing, and the definition of LLM-as-a-Judge has gradually taken on a clear and formal expression. A formal definition of LLM-as-a-Judge is as follows:

$$
P_{\theta}(X^{n}, C)\rightarrow R 
$$

\begin{itemize}
  \item \( P_{\theta} \): The LLM-as-a-Judge fulfilled by any LLM, which can be either foundation LLMs or fine-tuned version. The generation process is the auto-regressive process.
  \item \( X^{n} \): The samples to be evaluated. They can be of any available type, such as text, images, or videos. Here n represents the number of samples to be evaluated: When $n=1$, it becomes a point-wise judgment, where the evaluation result is a score. When $n=2$, it becomes a pair-wise judgment, where the evaluation result is a comparison. When $n > 2$, it becomes a list-wise judgment, where the evaluation result is presented as a ranking.
  \item \( C \): The context of the input \( x \), which includes relevant evaluation examples, historical information in the dialogue, or the definition of evaluation criteria.
  \item \( R \): The final evaluation result obtained from the LLM-as-a-Judge can be a relative or absolute score with rationale or not.
\end{itemize}

Based on the above definition, in this section, we introduce four strategies for making LLM-based evaluators more effective and robust: suitable prompt context, multi-evaluator collaboration, human-LLM collaboration, and better base LLMs.

\subsubsection{Suitable prompt context}
Careful prompt design is crucial for guiding an LLM judge to produce accurate and consistent evaluations. By tailoring the prompt with context, examples, or structured reasoning steps, researchers aim to align the model’s judgments with human criteria.
Below we review current methods that optimize the evaluation prompt from the perspectives of in-context samples, reasoning instruction, fine-grained criteria, and role-play augmentation.

\noindent \textbf{In-context Samples.} 
Demonstration plays an important role in the in-context learning. Many works~\cite{dong2024surveyincontextlearning,ye-etal-2023-complementary,xie2021explanation} have discussed the effectiveness regarding the sample selection, order, etc. Focusing on evaluation, a few high-quality demonstrations can calibrate the LLMs' expectations, guiding them in understanding and applying assessment standards.
Methods like GPTScore~\cite{fu2023gptscore} provide example answers with known quality to help the model learn how to assess text quality on the fly. 
Kotonya et al.~\cite{kotonya-etal-2023-little} shows the effectiveness of combining multiple prompt design methods with zero-shot and one-shot in-context samples, and the CoT prompt-based method shows considerable potential for assessing the quality of generated summaries.
Few-shot prompts make evaluation training-free and adaptable, but they can also introduce bias if the selected examples are unrepresentative.
To mitigate such bias, ALLURE~\cite{hasanbeig2023allure} iteratively refines the in-context examples by identifying erroneous evaluation outcomes, correcting them, and incorporating the revised results as updated examples. 
Alternatively, Song et al.~\cite{song2024can} introduce two types of many-shot in-context learning prompts, Many-Shot with Reference (MSwR) and Many-Shot without Reference (MSoR), to combat position or symbol biases.

\noindent \textbf{Reasoning instruction.} Evaluation also requires the reasoning ability to verify or infer the relationship between the response and the question. We roughly classify existing methods into two groups: CoT and planning instruction. A representative work in the first group is G-Eval~\cite{liu2023g}, which designs an auto-CoT framework that instructs LLMs to automatically generate evaluation steps given criteria before scoring. This framework with GPT-4 as the backbone model significantly improves the assessment of text summarization and dialogue generation tasks, achieving a high correlation with human evaluations.
To further evaluate the effectiveness of G-Eval, Chiang et al.~\cite{chiang-lee-2023-closer} examine how specific details in G-Eval's evaluation process influence the correlation between ratings provided by LLMs and those given by humans. Their findings indicate that the auto CoT used in G-Eval does not always enhance alignment with human ratings. However, they also find that prompting the LLM to explain its own ratings consistently improves the correlation between ChatGPT's evaluations and human judgments.
Domain-specific evaluators like ICE-Score~\cite{zhuo2024ice} for code generation go further. The prompt includes detailed evaluation steps, criteria, and task definitions, leading the LLM through a checklist (e.g. correctness, efficiency) when scoring code.
The second group of methods target the planning ability --- one can improve the evaluation performance by reducing the evaluation difficulty through task decomposition. \cite{saha-etal-2024-branch} proposes the Branch-Solve-Merge (BSM) framework to evaluate responses by dividing tasks into parallel sub-tasks, then solving each sub-task separately, and finally merging the results into an overall assessment. While, SocREval~\cite{he-etal-2024-socreval} introduces the Socratic method to leverage a sequence of probing questions to refine the reasoning instruction.

\noindent \textbf{Fine-grained criteria.} 
Another prompt strategy is to explicitly embed evaluation criteria or rubrics into the prompt, so the LLM judge assesses each aspect independently.
This criteria decomposition makes the evaluation more transparent and objective. For example, some users prefer to informative responses while others may expect concise answers.
Some researchers have explored fine-grained evaluations by indicating specific aspects 
(e.g., Fluency, Coherence, etc.)~\cite{li2023generative,bai2024benchmarking,yu2024kieval} and detailed rubrics~\cite{gao2023human}~\cite{kim2024prometheus} via in-context learning.
For instance, Jain et al.~\cite{jain2023multi} investigate the efficacy of LLMs as multi-dimensional evaluators: coherence, relevance, consistency, and fluency, each with two example scores. Their findings indicate that the prompt design strategies perform on par with traditional evaluation frameworks in text summarization tasks. 
Similarly, FineSurE~\cite{song2024finesure} exemplifies this by breaking summarization quality into dimensions like faithfulness, completeness, and conciseness; the LLM performs fact-checking and key fact alignment for each before outputting an overall judgment.
Furthermore, HD-EVAL~\cite{liu2024hd} enhances principle-driven prompting with hierarchical criteria. The authors first decompose the evaluation aspects using an LLM and assign scores to each sub-metric. Then, an aggregator combines these sub-metric scores into a total score, with human-labeled results used to train the aggregator. 
To investigate whether the evaluator can recognize and differentiate between various evaluation criteria, Hu et al.~\cite{hu2024llm} summarize and define an explicit hierarchical classification system consisting of 11 criteria. Using these criteria to test the evaluation capabilities of models, they identify that LLMs often confuse different criteria. To address this issue, they train the evaluator using clearly defined criteria to mitigate the potential confusion of different evaluation standards by LLMs.
More studies on tuning evaluators will be introduced later.

\noindent \textbf{Multi-turn \& role-play augmentation.} 
To better align with human judgment, recent methods have introduced multi-turn or role-play instructions. AutoCalibrate~\cite{liu2024calibrating} leverages a multi-stage prompt refinement process: the LLM is first prompted to draft initial evaluation criteria for a task, then revise them, and finally apply them.
Another approach is to give the LLM a specific role or persona~\cite{dong2024can} like ``You are a strict grammar teacher'' or ``You are a helpful peer reviewer''.
This can inject diverse evaluative perspectives and make the LLM more adaptable to different contexts. However, there are also some concerns: overly narrow roles or poorly chosen criteria can bias the evaluation. The goal of all these prompt-based techniques is to supply \textit{just enough contextual guidance} so that the LLM’s inherent knowledge is steered toward accurate judging, minimizing randomness or bias in its responses. 

\subsubsection{Multi-Evaluator Collaboration}
\label{sec:multieval}
Relying on the results from a single LLM judge may not be reliable due to the various biases inherent in LLMs. Typical biases include:

\begin{itemize}
  \item Position Bias. Position bias refers to the tendency of LLMs to favor answers based on their position in the response. This bias is common in various natural language processing tasks~\cite{bai2024benchmarking,aslanyan2019position} as well as in human decision-making processes~\cite{ramesh2021zero}. Even advanced LLMs like ChatGPT and GPT-4 encounter this issue when acting as evaluators~\cite{wang2023pandalm,zheng2023judging}.
  \item Knowledge Bias. Knowledge bias occurs when the pre-trained data fails to include certain essential tasks or introduces potentially harmful knowledge, which can undermine the generative performance of LLMs. In the evaluation scenario, this bias occurs when the knowledge required for evaluation tasks exceeds the scope of the LLM judge's training.
  \item Style Bias. Style bias in LLMs refers to the tendency to favor certain writing styles or tones due to the patterns in the pre-trained data. This bias can affect the LLM’s judgment, leading to assign higher scores to outputs that align with its preferred style, regardless of content quality.
  \item Format Bias. Format bias refers to the situation where a judge is fine-tuned without a reference but validated with a reference, or vice versa, resulting in a mismatched format. LLM judges perform poorly in these mismatched formats.
\end{itemize}

To overcome this limitation, several architectures and techniques use multiple LLMs~\cite{li2024mateval} (or multiple instances of an LLM~\cite{zhao2024auto,yu2024kieval}) that either cooperate or compete, and then combines their outputs.
The intuition is that aggregating multiple perspectives can cancel out individual errors or biases and lead to more reliable outcomes.
There are two main types of methods: cooperative approaches and aggregation approaches.

The first group is \textbf{aggregated multi-agent evaluation}, where models judge independently and their results are fused later. Representative works include Language-Model-as-an-Examiner~\cite{bai2024benchmarking}, which let a panel of LLMs generate probing questions about a candidate answer and then independently evaluate the answer, aggregating their scores. This peer-questioning plus voting mimics how a committee of examiners might each test a student with different questions, leading to a well-rounded evaluation. 
The benefit of voting ensembles is their simplicity and parallelizability. However, if all models share a blind spot, the ensemble won't fix it. Also, how to ensemble their results is critical in the final evaluation quality. 
1) Beyond simple voting, more sophisticated aggregation methods assign different weights or roles to each evaluator. One idea is to weight judges by their past agreement with humans. PRE~\cite{chu2024pre} conducts a ``qualification exam'' to select LLMs as reviewers, and then weights their ratings based on how well each aligns with human judgments.
Differently, PiCO~\cite{ning2024pico} treats the evaluation problem as a constrained optimization. Multiple models answer questions and evaluate each other's answers, and an algorithm finds weights for each model's opinion to maximize overall consistency within the group.
2) Apart from weighting by quality, we can also assign different evaluation criteria to different models. AIME~\cite{patel2024aime} gives each of several LLM judges a specific aspect to score (e.g. one model focuses only on factual accuracy, another only on fluency), and then concatenating or fusing these aspect-specific scores into an overall evaluation. 
Similarly, HD-Eval~\cite{liu2024hd} uses a panel of evaluators where each handles a hierarchically decomposed subset of criteria.
3) Advanced aggregation schemes also borrow from consensus algorithms. For example, Gao et al.~\cite{gao2024bayesian} applied Bayesian models to calibrate win rates when many LLM evaluators are voting, correcting biases in pairwise preference aggregation. 
Others construct a preference graph from multiple weak judges and then use graph algorithms to derive a final ranking that is more transitive-consistent~\cite{hu2024language}.

Another group of method is \textbf{cooperative multi-agent evaluation}, where multiple LLMs interact, sharing information or engaging in debate, to reach a consensus.
In these setups, each model might handle a different sub-task or provide feedback on others. For example, WideDeep~\cite{zhang2023wider} uses an architecture that lets models share information at a ``neuro-level'', effectively merging their intermediate representations to improve joint decision-making.
Other work borrows from human workflows. Xu et al.~\cite{xu2023towards} simulate an academic review process --- each agent drafts a solution, then reviews others' work and revises its own answer based on received critiques.
Similarly, ABSEval~\cite{liang2024abseval} assigns four distinct agent roles (answer synthesis, critique, execution, commonsense reasoning) that sequentially interact to evaluate an answer.
By role assignment in a collaborative workflow, the evaluators complement each other’s strengths (one agent might catch logical errors, another factual errors, etc.). However, a notable risk is ``groupthink''. If the models have similar biases or training backgrounds, their agreement may simply reinforce a shared bias rather than provide truly independent perspectives. 
Designing agent diversity (e.g. using different model architectures or prompt viewpoints) may be a potential solution. Therefore, another line of works is to have LLMs debate or challenge each other’s answers in a competitive fashion. In such frameworks, LLMs take on roles of debaters and a separate judge (which could itself be an LLM or an ensemble of LLMs) decides the winner of the debate.
An example work is Auto-Arena~\cite{zhao2024auto}, where candidate models engage in multi-round debates over a question, pointing out flaws in each other’s responses.
Extensions of this idea, like the MORE and SAMRE architectures~\cite{bandi2024adversarial}, involve multiple advocate agents and iterative rebuttal rounds, resembling a courtroom with opposing counsel and a verdict delivered after several back-and-forths. 
By contrast, decentralized debate structures let all models freely converse without a single controller. ChatEval~\cite{chan2023chateval} assigns diverse roles to multiple LLMs (e.g. one may emphasize precision, another creativity) and lets them discuss an open-ended question collectively.
Similarly, PRD~\cite{li2023prd} has models not only rank each other’s answers but also discuss them, which helped reduce biases like self-enhancement (where a model favors responses similar to itself) and positional bias (favoring the first presented answer).
Competitive debates tend to reveal flaws through contradiction and defense, leading to a more nuanced judgment. The challenge, however, is complexity and cost: multi-round debates consume more computation, and if not carefully orchestrated, the interactions could go in circles or become incoherent. Nonetheless, adversarial multi-LLM evaluation is a promising way to stress-test answers and achieve consensus closer to human critical analysis.

For efficiency, a variant of multi-evaluator systems is the cascade approach, where judges are arranged in tiers of increasing strength or cost. The idea is to use cheaper (or less powerful) models to handle easy evaluations and reserve expensive state-of-the-art models for the tricky cases, thereby optimizing resource use while maintaining accuracy. Jung et al.~\cite{jung2024trust} propose Cascaded Selective Evaluation, where a small judge model first evaluates; only if its confidence is low or a decision boundary is ambiguous, a larger model (like GPT-4) is called in.
Similarly, CascadedEval~\cite{huang2024empirical} combines open-source fine-tuned judges with proprietary models in a pipeline, leveraging the strengths of each. The fine-tuned judge handles routine cases and the proprietary model corrects its failures.
Such cascades illustrate a pragmatic collaboration between models of different caliber. The main challenge is designing a reliable gating mechanism to decide when to escalate to the next tier. If tuned well, cascaded systems can be both efficient and robust, effectively forming a safety net where the final tier (strongest model or even a human) only handles the most uncertain evaluations.

\subsubsection{Human-LLM Collaboration} 
Despite advances in automated evaluation, human insight remains essential, especially for open-ended tasks where nuanced understanding or ethical considerations are critical. Human–LLM collaboration frameworks aim at combining both merits: the efficiency of LLM judgments and the reliability of human oversight. A straightforward solution is \textbf{humans as verifier}. The LLM judge operates almost autonomously, but a human performs a final check or adjustment on its outputs~\cite{pan2024human,chiang-lee-2023-closer}.

Another line of work is \textbf{humans as assistant}. In this setup, LLMs generate initial judgments, which humans then refine before a final judgment is made. For example,
HMCEval~\cite{zhang-etal-2021-human} proposed a human-machine collaborative framework for dialogue evaluation. it optimizes evaluation reliability while minimizing human effort. Through a sample assignment approach, it reduces human involvement by half while maintaining 99\% evaluation accuracy, demonstrating a highly efficient solution for reliable dialogue evaluation.
While, CoEval~\cite{li2023collaborative} lets LLMs first generate task-specific evaluation metrics, which are then judged by humans for their usefulness. Afterward, carefully selected metrics are input to the evaluator to obtain evaluation results, which are further refined by humans.
EvalGen~\cite{shankar2024validates} tackles the problem of ``criteria drift'', where an LLM's tendency to unintentionally change its evaluation standards over many responses. In EvalGen, humans periodically provide feedback on the LLM's judging criteria, keeping them aligned over time. This iterative refinement of the evaluation rubric, driven by human judgment, was found to improve the consistency and fairness of long-running automated evaluations.
In addition to assisting in the generation of evaluation metrics, evaluators can also be employed to help generate test samples. Ribeiro and Lundberg~\cite{ribeiro-lundberg-2022-adaptive} introduce AdaTest, a human-LLM collaborative approach for automatically generating unit tests to identify and fix bugs in NLP models. AdaTest significantly improves bug detection efficiency, making users 5-10x more effective than traditional methods. Additionally, Rastogi et al.\cite{rastogi2023supporting} enhance the AdaTest auditing tool with human-AI collaboration, creating AdaTest++ to rigorously evaluate commercial language models like GPT-3 and Azure's sentiment analysis. Their tool leverages human strengths in sensemaking and hypothesis testing, effectively identifying a wide range of failure modes.
Another work by Wang et al.~\cite{wang2024large} introduced a calibration framework to correct known biases of LLM judges via human guidance.

The benefit of human-in-the-loop methods is a high assurance of quality: before any score is finalized, a person has vetted the process. Besides, Human–LLM collaboration can also be used to continuously improve the evaluator over time. By analyzing where the LLM’s judgments disagree with humans, developers can refine prompts or fine-tune the model. While, the downside is scalability --- it requires human labor for each evaluation or each batch of evaluations, so it may not be as fast or cheap as fully automated methods. Thus, these approaches are often more useful for high-stakes settings (e.g. medical evaluation) where accuracy outweighs speed. Or, in open-ended tasks where the ``ground truth'' is subjective and context-dependent, human guidance helps keep the automated judge aligned with social values and the specific goals of the evaluation.

\begin{table*}[htbp]
\centering
\caption{Comparisons of example evaluators in general or specific domains.}
\resizebox{\linewidth}{!}{
\begin{tabular}{l c c c c c l}
\toprule
\textbf{Method} & \textbf{Format} & \textbf{Critiques} & \textbf{Multi-rubrics} & \textbf{Data} & \textbf{Tuning} & \textbf{Example rubrics} \\ 
\midrule
\textbf{Shepherd} & pointwise & yes & Overall & Human & SFT & Error Analysis \\ 
\midrule
\textbf{Themis} & pointwise & yes & Multiple & GPT-4 & SFT & \makecell[l]{Cohesiveness; Likability; Clarity; \\ Length; Engagement; etc.} \\ 
\midrule
\textbf{PandaLM} & pairwise & yes & Multiple w/o ratings & GPT-3.5 & SFT & \makecell[l]{Relative Conciseness; Clarity; Comprehensiveness; \\ Formality; Adherence to Instructions; etc.} \\
\midrule
\textbf{JudgeLM} & pointwise \& pairwise & yes & Multiple w/o ratings & GPT-4 & SFT & \makecell[l]{Helpfulness; Relevance; Accuracy; \\ Level of Details of Responses} \\ 
\midrule
\textbf{AUTO-J} & pointwise \& pairwise & yes & Multiple w/o ratings & GPT-4 & SFT & \makecell[l]{Core Idea Capturing; Concise; Coverage; Harmlessness;\\ 
 Creativity; Engagement; Information Richness; etc.} \\ 
\midrule
\textbf{Prometheus} & pointwise \& pairwise & yes & Multiple & GPT-4 & SFT & Each sample is assigned a specific evaluation measure \\ 
\midrule
\textbf{TIGERScore} & pointwise & yes & Multiple w/o ratings & GPT-4 & SFT & \makecell[l]{Comprehension; Accuracy; Informativeness; Coherence; \\ Fact Consistency; Fluency; Accuracy; etc.} \\ 
\midrule
\textbf{CritiqueLLM} & pointwise \& pairwise & yes & Multiple & GPT-4 & SFT & \makecell[l]{Accuracy; User Satisfaction; Logical Coherence; \\ Creativity; Richness; Overall Score} \\ 
\midrule
\textbf{HALU-J} & pointwise & yes & Overall & GPT-4o & DPO & Hallucination \\
\midrule
\textbf{Safety-J} & pointwise & yes & Overall & Human \& GPT-4 & SFT & Safety \\ 
\bottomrule
\end{tabular}}
\label{tab:evaluators}
\end{table*}

\subsubsection{Better base LLMs} 
All of the aforementioned methods enhance the LLM's evaluation capability without modifying the LLM parameters themselves. In this section, we focus on training LLMs specially for evaluation usage.
There are general-purpose evaluators as well as domain-specific evaluators focusing on particular issues (e.g. safety compliance or factual accuracy).
Below, we survey these advances from the perspectives of evaluation data curation and tuning techniques, followed by highlighting some representative systems.
Finally, we outline key trends and trade-offs between fine-tuned evaluators and prompting-based evaluation.

\noindent \textbf{Evaluation data construction.} 
In Section~\ref{sec:autodata}, we have detailed automated dataset curation methods including compilation, derivation, and generation. Therefore, we will not repeat those methods and only focus on the construction of evaluation data.
The difference is that the training data for evaluators need to include LLMs' responses, which we classified into ``context generation'' in Section~\ref{sec:autodata_generation}.
To annotate these responses, there are mainly two types of approaches: manually-labeled and auto-synthetic.

To obtain manual labels, a straightforward solution to hire experts for annotation~\cite{wang2023shepherd}. This usually results in high-quality, nuanced feedback, but it is costly and slow to scale.
To lower the cost, some works leverage existing resources like online community feedback or crowdsourced annotations. 
Shepherd~\cite{wang2023shepherd} collects user feedback from two well-known communities: Stack Exchange and the Pushshift Reddit. They treated the title and subtitle of a post as a question, the top-level comments as answers, and the replies to these comments as critiques. The quality of these critiques can be evaluated based on the net upvotes and downvotes. Similarly, Vu et al.~\cite{vu2024foundational} curated a large and diverse set of over 100 quality assessment tasks, encompassing more than 5 million human evaluations from publicly released human feedback. 
Such community-sourced critiques provide diverse, real-world error examples. However, these relies written by human and the standards for their feedback may differ from benchmarking's needs.

For auto-synthetic data, a common strategy is to have a powerful model directly generate assessment scores given the LLM’s response.
For example, Auto-J~\cite{li2023generative} implement a ``divide-and-conquer'' approach, where GPT-4 generates two critiques for each response, which are then merged into a more thorough critique before providing a final rating.
Except for pointwise ratings, many works favor pairwise comparisons.
The above Auto-J~\cite{li2023generative} integrates both. For pairwise data, they provide two responses to the evaluator and ask them to identify the criteria where the evaluations differ between the two.
PandaLM~\cite{wang2023pandalm} re-formulates and completes the samples from Alpaca 52K as tuple (instruction, input, response1, response2), and ask GPT to generate output tuples (evaluation result, evaluation reason, reference response).
JudgeLM~\cite{zhu2023judgelm} adopts a similar way but leverages GPT-4 acted as a ``teacher'' judge.
Except for the data format (i.e., comparison or ratings), scoring rubrics are also critical. Prometheus~\cite{kim2023prometheus} ues GPT-4 to enhance a set seed of manual rubrics and generate more. These newly generated rubrics not only provide clearer standards, but also are more favorable for GPT judge, thereby high-quality evaluation data.
According to various rubrics, the corresponding error analysis or critiques provide additional interpretability.
TIGERScore~\cite{jiang2023tigerscore} builts a dataset called MetricInstruct with instruction prompts that ask for error analysis, where each entry includes a model output and a list of errors (with types and severity) as the label.

The above two methods each has its own merits. Often the best results come from combining human expertise with LLM generation. For example, InstructScore~\cite{xu-etal-2023-instructscore} and TIGERScore~\cite{jiang2023tigerscore} uses explicit human instruction (defining what to evaluate) together with GPT-4’s implicit knowledge to label data. 
While, Safety-J~\cite{liu2024safety} leverages human refine or review initial safety critiques.

\noindent \textbf{Tuning techniques.}
Given an evaluation dataset, the next step is to train the LLM to produce desired judgments or critiques via standard post-training techniques, e.g., supervised finetuning (SFT)~\cite{kim2023prometheus,saunders2022self,zhu2023judgelm,wang2023shepherd} or direct preference optimization (DPO)~\cite{wang2024halu,hu2024themis,wu2024meta}.
By contrast, traditional RL is less commonly used for training evaluators, since obtaining a numeric reward for a correct evaluation is not trivial --- one would need a ``meta-evaluator''. However, RLHF (human feedback) is introduced for improvements. For example, Safety-J~\cite{liu2024safety} employs an iterative preference learning loop, which uses its own critiques to perform meta-evaluation and then prefers revisions that improve its performance.
Over iterations, this is akin to the model reinforcing behaviors that lead to more accurate safety judgments. Of course, when letting the evaluator evaluate its own outputs and improve, the strategy is related to self-RL (e.g. ``Self-Refine'' and ``Self-Reward'' methods).

Complementary to the above tuning techniques, some tricks were introduced to make an evaluator be stable and unbiased in its judgments --- some irrelevant factors like the order in which answers are presented may be captured during tuning.
To mitigate positional bias, JudgeLM~\cite{zhu2023judgelm} conducted swap-augmentation (shuffling answer order), and used reference support/drop techniques to teach the judge to rely on content rather than position or formatting.
For robustness, GPT-4-based evaluators have been shown to exhibit variability if prompts are paraphrased\footnote{\url{https://eugeneyan.com/writing/llm-evaluators/}}.
To counter this, recent research~\cite{hu2024themis} generated many paraphrased instructions and fine-tuned models to give consistent preferences.

Each of the above models has a unique emphasis. Table~\ref{tab:evaluators} lists some representative evaluators including general and some specialized evaluators, e.g., Safety-J~\cite{liu2024safety} for safety judgement or Halu-J~\cite{wang2024halu} for assessing hallucinations, which prompt GPT-4 to generate multiple pieces of evidence for each instance as well as the final critique based on evidence.

\section{Open Challenges and Future Directions}
\label{sec:dis}
In this section, we discuss three core challenges that characterize the path towards generalizable evaluation in the era of LLMs.

\subsection{Challenges in Capability-Based Evaluation}
As LLMs unify various tasks and show human-like abilities, we conclude the transition from task-centric to capability-based evaluation. Section~\ref{sec:bench} provides a comprehensive survey of capability-based benchmarks including isolated and integrated evaluation. Based on them, we observe two core challenges.

\textbf{On one hand, how to achieve optimal balance between the efficiency and generalization of evaluation?} We can see that those comprehensive benchmarks face inherent scalability challenges. Unlike training data which benefits from scaling laws, benchmark expansion cannot indefinitely cover all desired competencies. Even for the agent-based evaluation, they also face their own efficiency-generalization trade-offs due to dependency on environment design.
The core challenge lies in selecting optimal task combinations that maximize evaluation efficiency while enabling reliable prediction of model full capability spectrum with limited test data. Preliminary solutions incorporating interpretability techniques like MUI~\cite{cao2025revisiting} have been proposed, but these represent only initial steps.

\textbf{On the other hand, should the evaluation focus on fine-grained capabilities or comprehensive integration?} Integrated datasets enable multi-dimensional analysis of model capabilities to identify strengths and weaknesses as guidance for training, yet they often overlook the tightly coupled nature of these competencies. While, agent-based evaluation naturally integrates multiple capabilities' testing via some environments. The high competency threshold for meaningful participation often excludes smaller models. Meanwhile, it typically lacks granular interpretability, thus offering little guidance for model optimization.

\subsection{Challenges in Automated Evaluation}
As LLM capabilities expand, creating suitable evaluation data by hand and judging the model's responses become a bottleneck. Automation promises to keep pace with rapid model progress and reduce our reliance on expensive human labeling.
For the transition from manual to automated evaluation, Section~\ref{sec:autoeval} provides a comprehensive survey of dataset curation and evaluators. Now, we discuss the core challenges from the two aspects.

Recent progress in automated benchmarking shows a striking dependence on LLMs for data curation, yet \textbf{generating harder, more diverse, and genuinely high-quality test data remains challenging}.
First, the quality ceiling of synthetic data is bounded by the current capability of the generator model. As one tries to raise difficulty --- longer contexts, more intricate reasoning, multimodal grounding --- the fidelity of LLM-generated content drops sharply. Although LLMs will keep improving, evaluation difficulty will rise in tandem, preserving the difficulty–quality trade-off. In practice, many researchers now resort to teacher–student distillation: crafting challenging prompts with a stronger ``teacher'' model and using them to train or benchmark smaller systems. Ultimately, however, a true closed loop of continuous model improvement demands generation techniques that surpass current capability ceiling, not merely mirror it.
Second, verifying the quality of synthetic data is itself non-trivial. If one needs an even stronger ``super-LLM'' to vet examples, the workflow becomes circular: how does one generate or validate data for that super-LLM? When existing human-curated corpora are exhausted, the field risks a ``chicken-and-egg'' bottleneck in which no component can improve without better data from the other.
Third, although ``diversity'' is widely acknowledged as crucial, there is no unified measurement or definition. Recent work has proposed counting domains, capability categories, or difficulty levels~\cite{muennighoff2025s1}, and Shypula \textit{et al.}~\cite{shypula2025evaluating} introduce a new metric for data diversity and quality. Yet we still lack a fine-grained formalism that links specific diversity dimensions to learning efficiency. An even deeper question is whether optimal diversity should be model-specific? The data needed to expose weaknesses in a retrieval-augmented LLM may differ from that required for a domain-specific LLM. Addressing these gaps will be pivotal for next-generation evaluation pipelines, which we will further discuss it later.

For evaluators, a straightforward question is \textbf{prompt-based or tuned evaluators}. Compared with prompt-based evaluation, tuned evaluators offer substantial advantages in cost and throughput, yet they cannot fully replace a strong foundation model when factual knowledge or sophisticated reasoning is required. When fine-tuning data are sparse or poorly curated, a tuned judge is prone to new biases, and its generalization seldom matches that of the underlying base model. Even so, tuned evaluators remain highly promising.

In specific, the current trend of evaluation data curation favors large-scale synthetic labeling using LLMs, sometimes combined with human annotations. This looks like knowledge distillation, tuned evaluators ultimately will be capped by the teacher model. But, through some well-designed tricks and human involvements, tuned evaluators can be more robust. Furthermore, some studies let an evaluator critique its own judgments and retrain on those critiques, forming a self-refinement loop that can yield ever-improving judges.
Thus, we vision a rise of specialized ``judge'' models that focus on particular concerns, e.g., safety, bias, factuality, reasoning~\cite{chen2025judgelrm}, or on domains such as math~\cite{wang2024Math-Shepherd}. 
These niche evaluators incorporate domain knowledge (retrieval for factuality, step-by-step solution checking for math) that a general judge might not possess. The trend suggests an ensemble of evaluators, each an expert in checking a certain aspect, could be used together for thoroughly evaluation.
Finally, \textbf{fine-grained, explainable judgments also attract increasing research attention}. This not only builds user trust but also transforms evaluation into a form of error analysis. It enables using the judgments to directly improve the generative model, thus closing the loop between evaluation and revision.

\subsection{Challenges in Generalizable Evaluation}

The core challenge in the era of LLMs is ensuring that our evaluation method keep up with the essentially unbounded capabilities of future LLMs. Traditional evaluation is bounded in the sense that it uses a fixed set of test examples and metrics, often reflecting the existing capabilities of LLMs. But actually LLMs are moving targets --- their abilities grow with scale and training, while evaluation can not be expanded infinitely considering the efficiency.
We are increasingly observing that an evaluation which a new model excels at might no longer be discriminative (the model ``outgrew'' the test), or conversely, a model might possess latent capabilities that the evaluation fails to reveal. This mis-match between what models can do and what we measure them on is widening. Thus, a core future direction is designing generalizable evaluations that anticipate and extrapolate to new model behaviors, rather than being one-off, static tests.

One aspect of this is \textbf{forecasting model capabilities} from the perspective of evaluation method. If we had reliable ways to predict how a model will perform on a broad range of tasks before actually testing it (or before the model even exists), we could design better benchmarks and safety checks proactively. Scaling laws can be regarded as a typical early work. They provide empirical relationships between model size, training compute, or data and performance, so that we may know the LLMs' future performance even during the early training stage. The BIG-Bench~\cite{srivastava2023beyond} was also motivated by extrapolating performance, tasks in BIG-Bench were chosen to be beyond the reach of smaller models, with the expectation that progress would be measurable as models scale up.
A recent work proposed to reduce the number of tasks by training a generic assessor for predictive performance~\cite{pacchiardi100}.
Indeed, its results indicated that some tasks improve smoothly with model size while others show discontinuous leaps at certain scales. Understanding these patterns (why some abilities suddenly ``activate'' at a threshold) is crucial for forecasting. If we can identify predictors in smaller models or early training phases that correlate with later emergent capabilities, we could flag potential breakthroughs in advance. 
These predictions, while not perfect, help benchmark designers include tasks that will remain challenging at the next generation of models, thereby ``future-proofing'' evaluations to some extent.
That is, generalizable evaluation cares about ``How predictable are LLM capabilities?'' We may use integrated ability and cross-scenario data to forecast performance of yet-unseen models.

Another direction of generalizable evaluation is dealing with the \textbf{inherent coverage problem} from the perspective of datasets. An LLM's possible behaviors are virtually infinite (suppose our ultimate target is AGI or ASI), but any test set is finite. How can we ensure that a finite evaluation set meaningfully probes the vast space of model competence? One idea is to focus on maximizing diversity and coverage with minimal data (as discussed above).
To do so, instead of one-test-set-for-all, future evaluations might be adaptive or model-specific. For example, an evaluator could iteratively find areas where the model's performance is problematic and add more tests there, until performance stabilizes. This resembles adaptive testing in education, where questions are chosen based on a student's previous answers to pinpoint their proficiency. In the LLM context, we may need to keep generating follow-up questions with suitable difficulty to map out the boundaries of its capabilities. If the model easily handles all math questions but struggles with certain logic puzzles, the system would concentrate evaluation on the latter to fully characterize the weakness. Some preliminary work in the safety domain in this direction includes adversarial testing and red teaming methods, where an auxiliary model or algorithm tries to find inputs that make the model fail.
Going forward, model-specific diversity could become standard~\cite{cao2025revisiting}. Each new model might be evaluated with a tailored set of stress tests chosen to cover its potential blind spots (as identified by prior models or preliminary runs). The goal is to achieve broad coverage (knowledge, reasoning, multi-modal, instruction following, safety, etc.) with as few test items as possible by targeting representative challenges rather than exhaustively enumerating trivial cases. This not only makes evaluation more efficient but also more generalizable: a well-chosen small test suite could predict performance on a much larger distribution of tasks because it captures the essential difficulties.

The third direction of generalizable evaluation seeks to \textbf{predict as-yet-uncovered abilities given limited testing sets} from the perspective of metric. An preliminary attempt is the Model Utilization Index (MUI)~\cite{cao2025revisiting}. MUI augments traditional, outcome-oriented scores by incorporating mechanism interpretability techniques, whereas classical metrics concern what result the model produced on a fixed test set, MUI additionally measures how much internal effort the model expended to obtain that result. Extensive experiments reveal an intuitive law: performance score is inversely correlated with MUI. The basic idea is when judging a human's overall proficiency we weigh both outcome and effort, where equal performance achieved with less effort (lower MUI) signals greater competence.
Nevertheless, this line of work remains constrained by the present limits of interpretability research. Neuron-localization methods, for example, have been criticized for imperfectly disentangling functional sub-skills, potentially undermining MUI’s precision. Sparse-Autoencoder (SAE) approaches, while more expressive, currently lack off-the-shelf generalizability; training a SAE for every new foundation model is prohibitively expensive. Moreover, both families of the above techniques require white-box access and are inapplicable to closed-source LLMs.
Despite these hurdles, the marriage of interpretability and evaluation presents a promising path forward. By looking inside the model we may transcend the intrinsic ceiling of finite test sets, inferring latent strengths or weaknesses that static outcome metrics miss. In short, explainable-aware metrics such as MUI demonstrate how one can ``see the whole from a part'', uncovering a model's true potential with limited external data.

The last intriguing direction of generalizable evaluation is \textbf{using a model's minor signals or reasoning traces to discover hidden capabilities or weaknesses}, probably from the perspective of evaluators. As LLMs increasingly can show their work (through CoT prompting, rationale outputs, or just the open-ended response itself), we have new data to judge what the model ``knows'' or where it falls short.
Some recent works focus on evaluating reasoning traces~\cite{chen2025xverify}.
Anthropic's study~\cite{chen2025reasoning} shows that CoT explanations are not always faithful. Although a reasoning model occasionally discloses which prompts or intermediate deductions it used, in most cases the verbalized CoT only partially reflects the model's actual computation. Even so, CoT monitoring remains valuable. Because unexpected behaviors,especially ones that unfold over several steps, often leave detectable artefacts in the trace, a fine-grained analysis can still surface hidden patterns. In other words, by inspecting the style, structure, or subtle irregularities in a model's explanation, evaluators can uncover clues about latent strengths or systematic flaws that would be invisible in a simple right/wrong score.
Such process-oriented evaluation does more than mark an answer incorrect; it reveals why it is wrong, and that diagnostic insight generalizes to many other inputs, not just the specific question posed. By treating the model's own explanations as data to be checked for factual alignment and logical validity, the boundary between outcome evaluation and process evaluation begins to blur. If the underlying reasoning process is demonstrably sound --- even on problems we did not explicitly test --- we gain confidence in the model's broader reliability.

\section{Conclusion}
LLMs are improving at a pace that outstrips conventional evaluation pipelines. In this survey, we mapped that tension onto two transitions and highlight the core limitation.
1) From tasks to capabilities, we re-organize benchmarks around five core abilities, knowledge, reasoning, instruction following, multi-modality and safety. This yet raises two open questions: Efficiency~vs.~generality and Granularity~vs.~integration.
2) From human-curated to LLM-automated evaluation. Automation is essential for keeping pace, but it introduces its own difficulties like generating harder, more diverse, high-quality data and tuning explainable, fine-grained LLM judges.
3) Toward generalizable evaluation, the core obstacle is a coverage gap: finite test sets cannot scale with unbounded model abilities. We thus discuss the potential directions including predictive evaluation, 
adaptive datasets, generalizable metrics, and see-the-whole-from-a-part evaluator.
Addressing these challenges demands a hybrid toolbox. Only by scaling our evaluations as aggressively as we scale our models can we ensure that performance claims remain meaningful, reliable and fair.
In the future, because the field evolves month-by-month, we will maintain a living repository\footnote{Benchmarks for core capabilities are at \url{https://github.com/ALEX-nlp/Benchmark-of-core-capabilities/tree/main}, and auto-evaluation methods are at \url{https://github.com/ALEX-nlp/Chapter3_Awesome_Paper_List}}, and warmly invite contributions that refine, correct or extend the taxonomy presented here.

\bibliographystyle{IEEEtran}
\bibliography{references}

\vfill

\end{document}